\documentclass[10pt,twocolumn,letterpaper]{article}

\usepackage{cvpr}              % To produce the CAMERA-READY version
\usepackage{graphicx}
\usepackage{amsmath}
\usepackage{amssymb}
\usepackage{booktabs}
\usepackage{bigstrut}
\usepackage{array}
\usepackage{colortbl}
\usepackage[T1]{fontenc}
\usepackage{wrapfig,lipsum}
\usepackage{booktabs}
\usepackage{bigstrut}
\usepackage{multirow}
\usepackage{placeins}
\usepackage{gensymb}
\usepackage{xcolor}
\usepackage{soul}
\usepackage{makecell}
\makeatletter  %https://github.com/MCG-NKU/CVPR_Template/issues/3
\@namedef{ver@everyshi.sty}{}
\makeatother
\usepackage{tikz}
\usepackage{yfonts}
\usepackage{graphicx}
\usepackage{import}
\usepackage{pifont} % \ding{55} script X:  http://ctan.org/pkg/pifont
\usepackage{appendix}
\usepackage[accsupp]{axessibility}  % Improves PDF readability for those with disabilities.
\usepackage{hhline}
\usepackage{stfloats}
\usepackage{setspace}
\usepackage{bigstrut}
\usepackage{atbegshi,picture}   
\usepackage{mathtools}
\usepackage{cite} 
\usepackage[font={footnotesize}]{caption}

\definecolor{cvprblue}{rgb}{0.21,0.49,0.74}
\usepackage[pagebackref,breaklinks,colorlinks,citecolor=cvprblue]{hyperref}

\usepackage[capitalize]{cleveref}
\crefname{section}{Sec.}{Secs.}
\Crefname{section}{Section}{Sections}
\Crefname{table}{Table}{Tables}
\crefname{table}{Tab.}{Tabs.}

\newcommand{\boldparagraphstart}[1]{\vspace{1pt}\noindent \textbf{#1}}
\newcommand{\xmark}{\text{\ding{55}}}  % pifont
\definecolor{darkgreen}{RGB}{0,127,0}
\definecolor{darkred}{RGB}{200,0,0}
\def\greencheckmark{\textcolor{darkgreen}{\checkmark}}
\def\redxmark{\textcolor{darkred}{\xmark}}

\DeclareMathOperator*{\argmaxB}{argmax}

\newcommand*{\affmark}[1][*]{\textsuperscript{#1}}
\newcommand{\thickline}[0]{\Xhline{3pt}}

\newenvironment{myitem}{\begin{list}{$\bullet$}
{\setlength{\itemsep}{-0pt}
\setlength{\topsep}{0pt}
\setlength{\labelwidth}{5pt}
\setlength{\leftmargin}{10pt}
\setlength{\parsep}{-0pt}
\setlength{\itemsep}{0pt}
\setlength{\partopsep}{0pt}}}%
{\end{list}}

\def\paperID{1007} % *** Enter the Paper ID here
\def\confName{CVPR}
\def\confYear{2024}

\author{Bowen Wen\affmark[] \quad Wei Yang\affmark[] \quad Jan Kautz\affmark[] \quad Stan Birchfield\affmark[]\\ 
\\
{
\affmark[]NVIDIA}
}

\begin{document}

\title{FoundationPose: Unified 6D Pose Estimation and 
Tracking of Novel Objects}

\setlength{\belowdisplayskip}{2pt} 
\setlength{\abovedisplayskip}{2pt} 
\setlength{\belowdisplayshortskip}{2pt} 
\setlength{\abovedisplayshortskip}{2pt}

\maketitle

%%%%%%%%% ABSTRACT
\vspace{-20pt}
\begin{abstract}
We present FoundationPose, a unified foundation model for 6D object pose estimation and tracking, supporting both model-based and model-free setups. Our approach can be instantly applied at test-time to a novel object without fine-tuning,  as long as its CAD model is given, or a small number of reference images are captured. 
Thanks to the unified framework, the  downstream pose estimation modules are the same in both setups, with a neural implicit representation used for efficient novel view synthesis when no CAD model is available.  
Strong generalizability is achieved via large-scale synthetic training, aided by a large language model (LLM), a novel transformer-based architecture, and contrastive learning formulation. 
Extensive evaluation on multiple public datasets involving challenging scenarios and objects indicate our unified approach outperforms existing  methods specialized for each task by a large margin. In addition, it even achieves comparable results to instance-level methods despite the reduced assumptions. Project page: \url{https://nvlabs.github.io/FoundationPose/}
\end{abstract}

%%%%%%%%% BODY TEXT
\vspace{-10pt}
\section{Introduction}

Computing the rigid 6D transformation from the object to the camera, also known as object pose estimation, is crucial for a variety of applications, such as robotic manipulation~\cite{kappler2018real,wen2022catgrasp,wen2022you} and mixed reality~\cite{marchand2015pose}. Classic methods~\cite{he2021ffb6d,he2020pvn3d,park2019pix2pose,wen2020robust,labbe2020cosypose} 
are known as \emph{instance-level} because they only work on the specific object instance determined at training time.  
Such methods usually require a textured CAD model for generating training data, and they cannot be applied to an unseen novel object at test time. 
While \emph{category-level methods}~\cite{wang2019normalized,tian2020shape,chen2020learning,zhang2022ssp,lee2023tta} remove these assumptions (instance-wise training and CAD models), they are limited to objects within the predefined category on which they are trained.  Moreover, obtaining category-level training data is notoriously difficult, in part due to additional pose canonicalization and examination steps~\cite{wang2019normalized} that must be applied.

\begin{figure}[t]
    \centering
    {\includegraphics[width=0.4\textwidth]{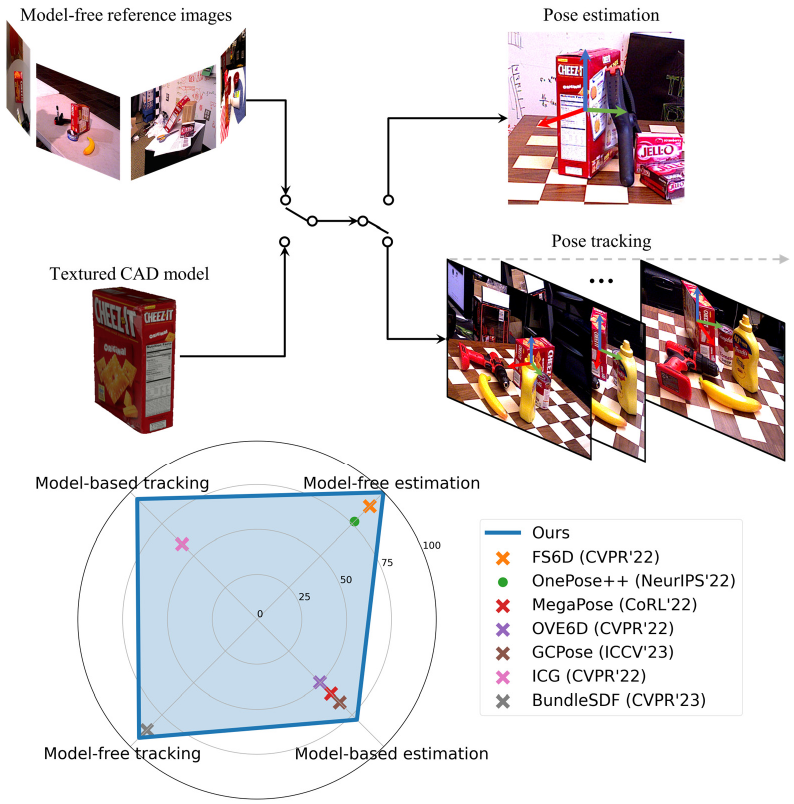}} 
    \vspace{-10pt}
    \caption{Our unified framework enables both 6D pose estimation and tracking for novel objects, supporting the model-based and model-free setups. On each of these four tasks, it outperforms prior work specially designed for the task ($\bullet$ indicates RGB-only; $\times$ indicates RGBD, like ours).  The metric for each task is explained in detail in the experimental results.}  
    \label{fig:intro}
    \vspace{-20pt}
\end{figure}

To address these limitations, more recent efforts have focused on the problem of instant pose estimation of arbitrary novel objects~\cite{shugurov2022osop,labbemegapose,sun2022onepose,he2022oneposepp,liu2022gen6d}. 
Two different setups are considered, depending upon what information is available at test time: \emph{model-based}, where a textured 3D CAD model of the object is provided, and \emph{model-free}, where a set of reference images of the object is provided. 
While much progress has been made on both setups individually, there remains a need for a single method to address both setups in a unified way, since different real-world applications provide different types of information.  

Orthogonal to single-frame object pose estimation, pose tracking methods~\cite{wen2020se,stoiber2022iterative,deng2019pose,li2018deepim,wang20206,lin2022keypoint,wuthrich2013probabilistic,issac2016depth} leverage temporal cues to enable more efficient, smooth and accurate pose estimation on a video sequence. These methods share the similar aforementioned issues to their counterparts in pose estimation, depending on their assumptions on the object knowledge.

In this paper we propose a unified framework called \textit{FoundationPose} that performs both pose estimation and tracking for novel objects in both the model-based and model-free setups, using RGBD images. As seen in Fig.~\ref{fig:intro}, our method outperforms existing state-of-art methods specialized for each of these four tasks. Our strong generalizability is achieved via large-scale synthetic training, aided by a large language model (LLM), as well as a novel transformer-based architecture and contrastive learning.  
We bridge the gap between model-based and model-free setups with a neural implicit representation that allows for effective novel view synthesis with a small number ($\sim$16) of reference images, achieving rendering speeds that are significantly faster than previous render-and-compare methods~\cite{li2018deepim,labbemegapose,wen2020se}. 
Our contributions can be summarized as follows:
\begin{myitem}
    \item We present a unified framework for both pose estimation and tracking for novel objects, supporting both model-based and model-free setups. An object-centric neural implicit representation for effective novel view synthesis bridges the gap between the two setups.
    \item We propose a LLM-aided synthetic data generation pipeline which scales up the variety of 3D training assets by diverse texture augmentation.
    \item Our novel design of transformer-based network architectures and contrastive learning formulation leads to strong generalization when trained solely on synthetic data.
    \item Our method outperforms existing  methods specialized for each task by a large margin across multiple public datasets. It even achieves comparable results to instance-level methods despite reduced assumptions.
\end{myitem}
Code and data developed in this work will be released.
%to benefit the community.

\section{Related Work}

\boldparagraphstart{CAD Model-based Object Pose Estimation.} Instance-level pose estimation methods~\cite{he2021ffb6d,he2020pvn3d,park2019pix2pose,labbe2020cosypose} assume a textured CAD model is given for the object. Training and testing is performed on the exact same instance. The object pose is often solved by direct regression~\cite{xiang2018posecnn,li2019coordinates}, or constructing 2D-3D correspondences followed by P$n$P~\cite{tremblay2018deep,park2019pix2pose}, or 3D-3D correspondences followed by least squares fitting~\cite{he2021ffb6d,he2020pvn3d}. To relax the assumptions about the object knowledge, category-level methods~\cite{wang2019normalized,tian2020shape,chen2020learning,zhang2022ssp,lee2023tta,zheng2023hs} can be applied to novel object instances of the same category, but they cannot generalize to arbitrary novel objects beyond the predefined categories. To address this limitation, recent efforts~\cite{labbemegapose,shugurov2022osop} aim for instant pose estimation of arbitrary novel objects as long as the CAD model is provided at test time. 

\boldparagraphstart{Few-shot Model-free Object pose estimation.} Model-free methods remove the requirement of an explicit textured model. Instead, a number of reference images capturing the target object are provided~\cite{park2020latentfusion,he2022fs6d,sun2022onepose,he2022oneposepp}. RLLG~\cite{cai2020reconstruct} and NeRF-Pose~\cite{li2023nerf} propose instance-wise training without the need of an object CAD model. In particular, \cite{li2023nerf} constructs  a neural radiance field to provide semi-supervision on the object coordinate map and mask. 
Differently, we introduce the neural object field built on top of SDF representation for efficient RGB and depth rendering to bridge the gap between the model-based and model-free scenarios. In addition, we focus on generalizable novel object pose estimation in this work, which is not the case for \cite{cai2020reconstruct,li2023nerf}. To handle novel objects, Gen6D~\cite{liu2022gen6d} designs a detection, retrieval and refinement pipeline. However, to avoid difficulties with out-of-distribution test set, it requires fine-tuning. OnePose~\cite{sun2022onepose} and its extension OnePose++~\cite{he2022oneposepp} leverage structure-from-motion (SfM) for object modeling and pretrain 2D-3D matching networks to solve the pose from correspondences. FS6D~\cite{he2022fs6d} adopts a similar scheme and focuses on RGBD modality. Nevertheless, reliance on correspondences becomes fragile when applied to textureless objects or under severe occlusion.

\boldparagraphstart{Object Pose Tracking.} 
6D object pose  tracking aims to leverage temporal cues to enable more efficient, smooth and accurate pose prediction on video sequence. Through neural rendering, our method can be trivially extended to the pose tracking task with high efficiency. Similar to single-frame pose estimation, existing tracking methods can be categorized into their counterparts depending on the assumptions of object knowledge. These include instance-level methods~\cite{wen2020se,li2018deepim,deng2019pose,garon2018framework}, category-level methods~\cite{wang20206,lin2022keypoint},  model-based novel object tracking~\cite{stoiber2022iterative,wuthrich2013probabilistic,issac2016depth} and model-free novel object tracking~\cite{bundle2021wen,wen2023bundlesdf}. 
Under both model-based and model-free setups, we set a new benchmark record across public datasets, even outperforming state-of-art methods that require instance-level training~\cite{wen2020se,li2018deepim,deng2019pose}.

%%%%%%%%%%%%%%%%%%%%%%%
\section{Approach}

\begin{figure*}[h]
    \vspace{-12pt}
    \centering
    {\includegraphics[width=0.98\textwidth]{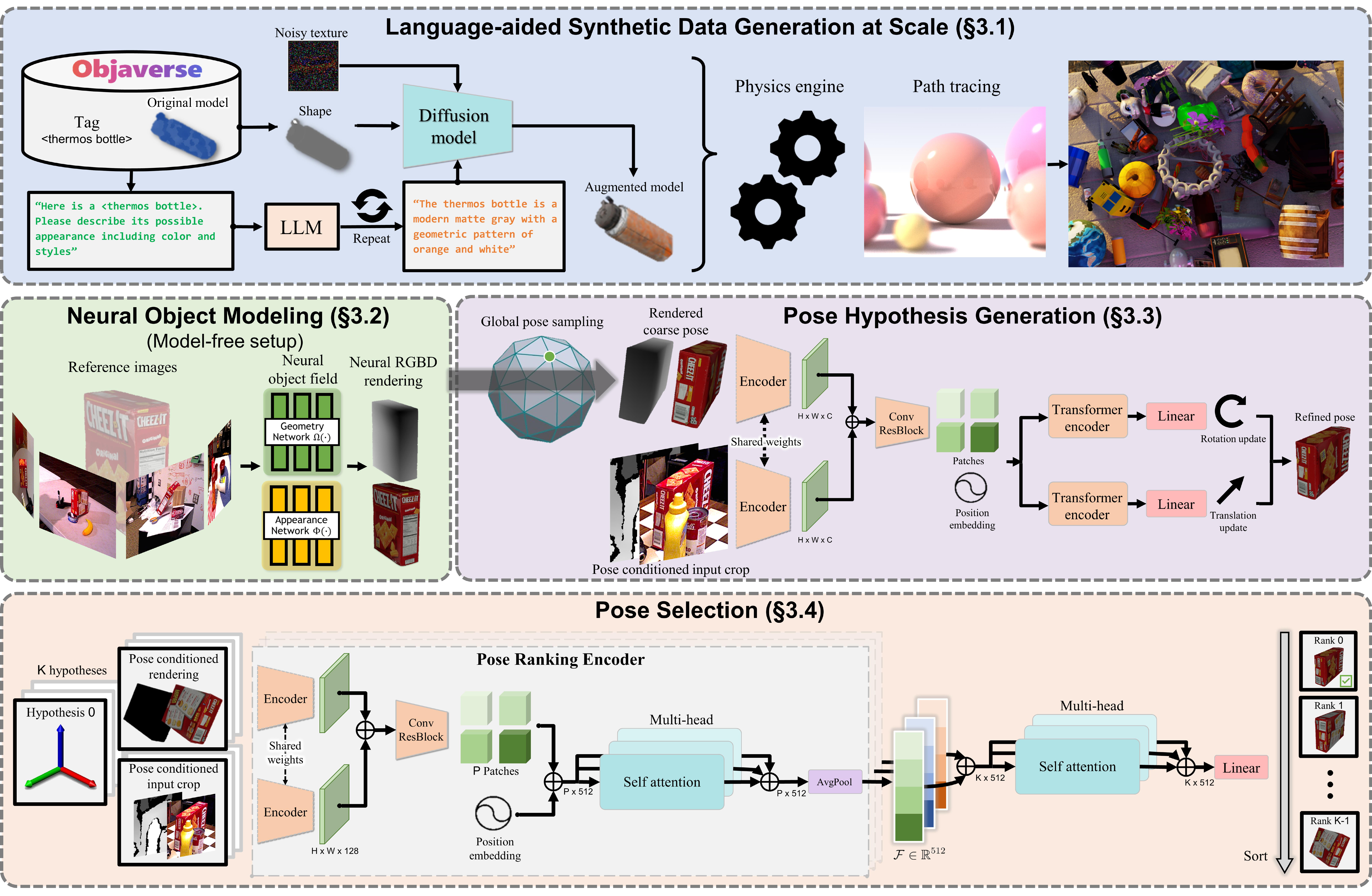}} 
    \vspace{-12pt}
    \caption{Overview of our framework. To reduce manual efforts for large scale training, we developed a novel synthetic data generation pipeline by leveraging recent emerging techniques and resources including 3D model database, large language models and diffusion models (Sec.~\ref{sec:language}). To bridge the gap between model-free and model-based setup, we leverage an object-centric neural field (Sec.~\ref{sec:nerf}) for novel view RGBD rendering for subsequent render-and-compare. For pose estimation, we first initialize global poses uniformly around the object, which are then refined by the refinement network (Sec.~\ref{sec:refiner}). Finally, we forward the refined poses to the pose selection module which predicts their scores. The pose with the best score is selected as output (Sec.~\ref{sec:ranking}).}\label{fig:pipeline}
    \vspace{-0.2in}
\end{figure*}

Our system as a whole is illustrated in Fig.~\ref{fig:pipeline}, showing the relationships between the various components, which are described in the following subsections.

%Our method is described in the following subsections.  The relationships between the subsections, and the system as a whole, are illustrated in Fig.~\ref{fig:pipeline}.

\subsection{Language-aided Data Generation at Scale}\label{sec:language}
To achieve strong generalization, a large diversity of objects and scenes is needed for training. Obtaining such data in the real world, and annotating accurate ground-truth 6D pose, is time- and cost-prohibitive.  Synthetic data, on the other hand, often lacks the size and diversity in 3D assets.  We developed a novel synthetic data generation pipeline for training, powered by the recent emerging resources and techniques: large scale 3D model database~\cite{deitke2023objaverse,downs2022google}, large language models (LLM), and diffusion models~\cite{rombach2022high,ho2020denoising,cao2023texfusion}. This approach dramatically scales up both the amount and diversity of data compared with prior work~\cite{hodan2018bop,he2022fs6d,labbemegapose}.   

\boldparagraphstart{3D Assets.} 
 We obtain training assets from recent large scale 3D databases including Objaverse~\cite{deitke2023objaverse} and GSO~\cite{downs2022google}. For Objaverse~\cite{deitke2023objaverse} we chose the objects from the Objaverse-LVIS subset that consists of more than 40K objects belonging to 1156 LVIS~\cite{gupta2019lvis} categories. 
This list contains the most relevant daily-life objects with reasonable quality, and diversity of shapes and appearances. It  also provides a tag for each object describing its category, which benefits automatic language prompt generation in the following LLM-aided texture augmentation step.

\begin{figure}[t]
    \centering
    \vspace{-0.1in}
    {\includegraphics[width=0.4\textwidth]{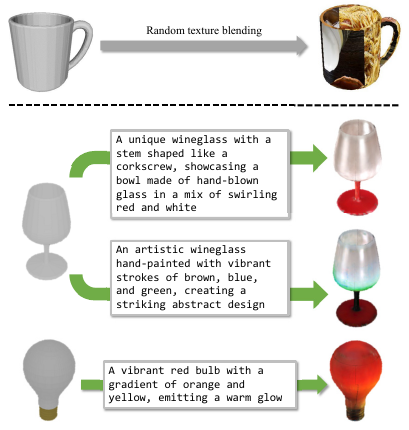}} 
    \vspace{-0.1in}
    \caption{\textbf{Top:} Random texture blending proposed in FS6D~\cite{he2022fs6d}. \textbf{Bottom:} Our LLM-aided texture augmentation yields more realistic appearance. Leftmost is the original 3D assets. Text prompts are automatically generated by ChatGPT.}\label{fig:tex_aug}
    \vspace{-20pt}
\end{figure}

\boldparagraphstart{LLM-aided Texture Augmentation.} While most Objaverse objects have high quality shapes, their texture fidelity varies significantly. FS6D~\cite{he2022fs6d} proposes to augment object texture by randomly pasting images from ImageNet~\cite{deng2009imagenet} or MS-COCO~\cite{lin2014microsoft}. However, due to the random UV mapping, this method yields artifacts such as seams on the resulting textured mesh (Fig.~\ref{fig:tex_aug} top); and applying holistic scene images to objects leads to unrealistic results. In contrast, we explore how recent advances in large language models and diffusion models can be harnessed for more realistic (and fully automatic) texture augmentation. Specifically, we provide a text prompt, an object shape, and a randomly initialized noisy texture to TexFusion~\cite{cao2023texfusion} to produce an augmented textured model. Of course, providing such a prompt manually is not scalable if we want to augment a large number of objects in diverse styles under different prompt guidance. As a result, we introduce a two-level hierarchical prompt strategy. As illustrated in Fig.~\ref{fig:pipeline} top-left, we first prompt ChatGPT, asking it to describe the possible appearance of an object; this prompt is templated so that each time we only need to replace the tag paired with the object, which is given by the Objaverse-LVIS list. The answer from ChatGPT then becomes the text prompt provided to the diffusion model for texture synthesis. Because this approach enables full automation for texture augmentation, it facilitates diversified data generation at scale. Fig.~\ref{fig:tex_aug} presents more examples including different stylization for the same object.

\boldparagraphstart{Data Generation.} Our synthetic data generation is implemented in NVIDIA Isaac Sim, leveraging path tracing for high-fidelity photo-realistic  rendering.\footnote{\url{https://developer.nvidia.com/isaac-sim}}  We perform gravity and physics simulation to produce physically plausible scenes. In each scene, we randomly sample objects including the original and texture-augmented versions. The object size, material, camera pose, and lighting are also randomized; more details can be found in the appendix.

\subsection{Neural Object Modeling}\label{sec:nerf}

For the model-free setup, when the 3D CAD model is unavailable, one key challenge is to represent the object to effectively render images with sufficient quality for downstream modules. 
Neural implicit representations are both effective for novel view synthesis and parallelizable on a GPU, thus providing high computational efficiency when rendering multiple pose hypotheses for downstream pose estimation modules, as shown in Fig.~\ref{fig:pipeline}.
To this end, we introduce an object-centric neural field representation for object modeling, inspired by previous work~\cite{wang2021neus,mueller2022instant,yariv2020multiview,wen2023bundlesdf}.

\boldparagraphstart{Field Representation.} We represent the object by two functions~\cite{yariv2020multiview} as shown in Fig.~\ref{fig:pipeline}.  First, the geometry function $\Omega: x \mapsto  s$ takes as input a 3D point $x \in \mathbb{R}^3$ and outputs a signed distance value $s \in \mathbb{R}$.  Second, the appearance function $\Phi: (f_{\Omega(x)}, n, d) \mapsto c$ takes the intermediate feature vector $f_{\Omega(x)}$ from the geometry network, a point normal $n \in \mathbb{R}^3$, and a view direction $d \in \mathbb{R}^3$, and outputs the color  $c \in \mathbb{R}^3_+$. 
In practice, we apply multi-resolution hash encoding \cite{mueller2022instant} to $x$ before forwarding to the network. 
Both $n$ and $d$ are embedded by a fixed set of second-order spherical harmonic coefficients. The implicit object surface is obtained by taking the zero level set of the signed distance field (SDF): 
$S=\left \{ x\in \mathbb{R}^3 \mid \Omega(x)=0  \right \}$. Compared to NeRF~\cite{mildenhall2021nerf}, the SDF representation $\Omega$ provides higher quality depth rendering while removing the need to manually select a density threshold.

\boldparagraphstart{Field Learning.} 
For texture learning, we follow the volumetric rendering over truncated near-surface regions~\cite{wen2023bundlesdf}:
\begin{align}
    &c(r)=\int_{z(r)-\lambda}^{z(r)+0.5\lambda} w(x_i)\Phi(f_{\Omega(x_i)},n(x_i),d(x_i))\,dt, \label{eq:render} \\
    &w(x_i)= \frac{1}{1+e^{-\alpha\Omega(x_i)}}\frac{1}{1+e^{\alpha\Omega(x_i)}},
\end{align}
where $w(x_i)$ is the bell-shaped probability density function \cite{wang2021neus} that depends on the signed distance $\Omega(x_i)$ from the point to the implicit object surface, and $\alpha$ adjusts the softness of the distribution. The probability peaks at the surface intersection.  In Eq.~\eqref{eq:render}, $z(r)$ is the depth value of the ray from the depth image, and $\lambda$ is the truncation distance.  We ignore the contribution from empty space that is more than $\lambda$ away from the surface for more efficient training, and we only integrate up to a $0.5\lambda$ penetrating distance to model self-occlusion \cite{wang2021neus}. During training, we compare this quantity against the reference RGB images for color supervision:
\begin{align}
    \mathcal{L}_{c}=\frac{1}{|\mathcal{R}|}\sum_{r\in \mathcal{R}}\left \| c(r)-\bar{c}(r) \right \|_2,
\end{align}
where $\bar{c}(r)$ denotes the ground-truth color at the pixel where the ray $r$ passes through. 

For geometry learning, we adopt the hybrid SDF model~\cite{wen2023bundlesdf} by dividing the space into two regions to learn the SDF, leading to the empty space loss and the near-surface loss. We also apply eikonal regularization~\cite{gropp2020implicit} to the near-surface SDF: 
\begin{align}
&\mathcal{L}_{\textit{e}}=\frac{1}{|\mathcal{X}_{\textit{e}}|}\sum_{x\in \mathcal{X}_{\textit{e}}} | \Omega(x)-\lambda |, \\
&\mathcal{L}_{\textit{s}}=\frac{1}{|\mathcal{X}_{\textit{s}}|}\sum_{x\in \mathcal{X}_{\textit{s}}}\left(\Omega(x)
    +d_x - d_D \right)^2, \\
&\mathcal{L}_{\textit{eik}}=\frac{1}{|\mathcal{X}_{\textit{s}}|}\sum_{x\in \mathcal{X}_{\textit{s}}} ( \left\|\nabla  \Omega(x)\right\|_2-1 )^2,
\end{align}
where $x$ denotes a sampled 3D point along the rays in the divided space;  $d_x$ and $d_D$ are the distance from ray origin to the sample point and the observed depth point, respectively.  We do not use the uncertain free-space loss~\cite{wen2023bundlesdf}, as the template images are pre-captured offline in the model-free setup.
The total training loss is
\begin{equation}
\begin{aligned}
\mathcal{L}=w_{c}\mathcal{L}_{c}+w_{\textit{e}}\mathcal{L}_{\textit{e}}+w_{\textit{s}}\mathcal{L}_{\textit{s}}+w_{\textit{eik}}\mathcal{L}_{\textit{eik}}.
\end{aligned}
\end{equation}
The learning is optimized per object without priors and can be efficiently performed within seconds. 
The neural field only needs to be trained once for a novel object.  

\boldparagraphstart{Rendering.} Once trained, the neural field can be used as a drop-in replacement for a conventional graphics pipeline, to perform efficient rendering of the object for subsequent render-and-compare iterations. In addition to the color rendering as in the original NeRF~\cite{mildenhall2021nerf}, we also need depth rendering for our RGBD based pose estimation and tracking. To do so, we perform marching cubes~\cite{lorensen1998marching} to extract a textured mesh from the zero level set of the SDF, combined with color projection. This only needs to be performed once for each object. At inference, given an object pose, we then render the RGBD image following the rasterization process. Alternatively, one could directly render the depth image using $\Omega$ online with sphere tracing~\cite{hart1996sphere}; however, we found this leads to less efficiency, especially when there is a large number of pose hypotheses to render in parallel.

\subsection{Pose Hypothesis Generation}\label{sec:refiner}
\boldparagraphstart{Pose Initialization.} Given the RGBD image, the object is detected using an off-the-shelf method such as Mask R-CNN~\cite{he2017mask} or CNOS~\cite{nguyen2023cnos}.  We initialize the translation using the 3D point located at the median depth within the detected 2D bounding box. 
% \begin{equation}
%     t^{\text{init}}=z_m\pi^{-1}[u_m, v_m, 1]^{\top}
% \end{equation}
To initialize rotations, we uniformly sample $N_s$ viewpoints from an icosphere centered on the object with the camera facing the center. These camera poses are further augmented with $N_i$ discretized in-plane rotations, resulting in $N_s\cdot N_i$ global pose initializations which are sent as input to the pose refiner.

\boldparagraphstart{Pose Refinement.} Since the coarse pose initializations from the previous step are often quite noisy, a refinement module is needed to improve the pose quality. Specifically, we build a pose refinement network which takes as input the rendering of the object conditioned on the coarse pose, and a crop of the input observation from the camera; the network outputs a pose update that improves the pose quality. Unlike MegaPose~\cite{labbemegapose}, which renders multiple views around the coarse pose to find the anchor point, we observed rendering a single view corresponding to the coarse pose suffices. For the input observation, instead of cropping based on the 2D detection which is constant, we perform a pose-conditioned cropping strategy so as to provide feedback to the translation update. Concretely, we project the object origin to the image space to determine the crop center. We then project the slightly enlarged object diameter (the maximum distance between any pair of points on the object surface) to determine the crop size that encloses the object and the nearby context around the pose hypothesis. This crop is thus conditioned on the coarse pose and encourages the network to update the translation to make the crop better aligned with the observation.  The refinement process can be repeated multiple times by feeding the latest updated pose as input to the next inference, so as to iteratively improve the pose quality.

The refinement network architecture is illustrated in Fig.~\ref{fig:pipeline}; details are in the appendix. We first extract feature maps from the two RGBD input branches with a single shared CNN encoder. The feature maps are concatenated, fed into CNN blocks with residual connection~\cite{he2016deep}, and tokenized by dividing into patches~\cite{dosovitskiy2020image} with position embedding. Finally, the network predicts the translation update $\Delta \boldsymbol{t}\in \mathbb{R}^{3}$ and rotation update $\Delta \boldsymbol{R}\in \mathbb{SO}(3)$, each individually processed by a transformer encoder~\cite{vaswani2017attention} and linearly projected to the output dimension. More concretely, $\Delta \boldsymbol{t}$ represents the object's translation shift in the camera frame, $\Delta \boldsymbol{R}$ represents the object's orientation update expressed in the camera frame. In practice, the rotations are parameterized in axis-angle representation. We also experimented with the 6D representation~\cite{zhou2019continuity} which achieves similar results. The input coarse pose $[\boldsymbol{R} \,|\, \boldsymbol{t}] \in \mathbb{SE}(3)$ is then updated by:
\begin{equation}
    \boldsymbol{t}^{+}=\boldsymbol{t}+\Delta \boldsymbol{t}
\end{equation}
\vspace{-0.2in}
\begin{equation}
    \boldsymbol{R}^{+}=\Delta\boldsymbol{R}\otimes \boldsymbol{R},
\end{equation}
where $\otimes$ denotes update on $\mathbb{SO}(3)$. Instead of using a single homogeneous pose update, 
% this disentangled representation removes the dependency of orientation upon translation, and vice versa, thus simplifying the learning process.
this disentangled representation removes the dependency on the updated orientation when applying the translation update. This unifies both the updates and input observation in the camera coordinate frame and thus simplifies the learning process. 
The network training is supervised by $L_2$ loss:
\begin{equation}\label{eq:refine_loss}
    \mathcal{L}_{\text{refine}}=w_1\left \| \Delta\boldsymbol{t}-\Delta\bar{\boldsymbol{t}} \right \|_2 + w_2\left \| \Delta\boldsymbol{R}-\Delta\bar{\boldsymbol{R}} \right \|_2,
\end{equation}
where $\bar{\boldsymbol{t}}$ and $\bar{\boldsymbol{R}}$ are ground truth; $w_1$ and $w_2$ are the weights balancing the losses, which are set to 1 empirically.

\subsection{Pose Selection}\label{sec:ranking}
Given a list of refined pose hypotheses, we use a hierarchical pose ranking network to compute their scores. The pose with the highest score is selected as the final estimate.

\begin{figure}[t]
    % \vspace{-10pt}
    \centering
    {\includegraphics[width=0.47\textwidth]{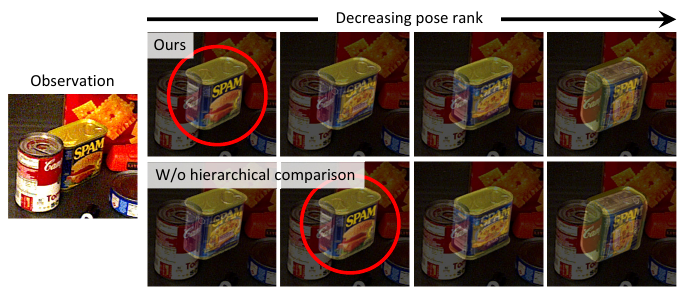}} 
    \vspace{-10pt}
    \caption{Pose ranking visualization. Our proposed hierarchical comparison leverages the global context among all pose hypotheses for a better overall trend prediction that aligns both shape and texture. The true best pose is annotated with red circle.} \label{fig:pose_rank}    % Stan:  The word 'optimal' is dangerous
    \vspace{-0.2in}
\end{figure}

\boldparagraphstart{Hierarchical Comparison.} The network uses a two-level comparison strategy. First, for each pose hypothesis, the rendered image is compared against the cropped input observation, using the pose-conditioned cropping operation was introduced in Sec.~\ref{sec:refiner}. This comparison (Fig.~\ref{fig:pipeline} bottom-left) is performed with a pose ranking encoder, utilizing the same backbone architecture for feature extraction as in the refinement network. The extracted features are concatenated, tokenized and forwarded to the multi-head self-attention module so as to better leverage the global image context for comparison. The pose ranking encoder performs average pooling to output a feature embedding $\mathcal{F}\in \mathbb{R}^{512}$ describing the alignment quality between the rendering and the observation (Fig.~\ref{fig:pipeline} bottom-middle). At this point, we could directly project $\mathcal{F}$ to a similarity scalar as typically done~\cite{nguyen2022templates,labbemegapose,cai2022ove6d}. However, this would ignore the other pose hypotheses, forcing the network to output an absolute score assignment which can be difficult to learn. 

To leverage the global context of all pose hypotheses in order to make a more informed decision, we introduce a second level of comparison among all the $K$ pose hypotheses.
Multi-head self-attention is performed on the concatenated feature embedding $\mathbf{F}=[\mathcal{F}_0, \ldots, \mathcal{F}_{K-1}]^{\top} \in \mathbb{R}^{K\times 512}$, which encodes the pose alignment information from all poses. 
By treating $\mathbf{F}$ as a sequence, this approach naturally generalizes to varying lengths of $K$~\cite{vaswani2017attention}. We do not apply position encoding to $\mathbf{F}$, so as to be agnostic to the permutation. The attended feature is then linearly projected to the scores $\mathbf{S} \in \mathbb{R}^K$ to be assigned to the pose hypotheses. The effectiveness of this hierarchical comparison strategy is shown in a typical example in Fig.~\ref{fig:pose_rank}.

\boldparagraphstart{Contrast Validation.} To train the pose ranking network, we propose a \textit{pose-conditioned triplet loss}:
\begin{equation}\label{eq:triplet}
    \mathcal{L}(i^+, i^-)=\text{max}(\mathbf{S}(i^{-})-\mathbf{S}(i^{+})+\alpha, 0),
\end{equation}
where $\alpha$ denotes the contrastive margin; $i^{-}$ and $i^{+}$ represent the negative and positive pose samples, respectively, which are determined by computing the ADD metric~\cite{xiang2018posecnn} using ground truth. Note that different from standard triplet loss~\cite{hoffer2015deep}, the anchor sample is not shared between the positive and negative samples in our case, since the input is cropped depending on each pose hypothesis to account for translations. 
While we can compute this loss over each pair in the list, the comparison becomes ambiguous when both poses are far from ground truth.
Therefore, we only keep those pose pairs whose positive sample is from a viewpoint that is close enough to the ground truth to make the comparison meaningful:
\begin{align}
    \mathbb{V}^+ &=\{i \,:\, D(\boldsymbol{R}_i, \bar{\boldsymbol{R}})<d \} \\
    \mathbb{V}^- &= \{0, 1, 2, \ldots, K-1\} \\ %\mathbb{Z}_K \\ 
    \mathcal{L}_{\text{rank}} &=\sum_{i^+, i^-} \mathcal{L}(i^+, i^-)
    % \mathcal{L}_{\text{rank}} &=\sum_{i^+\in \mathbb{V},i^-\in|K|}^{} \mathcal{L}(i^+, i^-)
   \label{eq:list_loss}
\end{align}
where the summation is over $i^+\in \mathbb{V}^+,i^-\in \mathbb{V}^-, i^+ \neq i^-$; $\boldsymbol{R}_i$ and $\bar{\boldsymbol{R}}$ are the rotation of the hypothesis and ground truth, respectively; $D(\cdot)$ denotes the geodesic distance between rotations; and $d$ is a predefined threshold. We also experimented with the InfoNCE loss~\cite{oord2018representation} as used in \cite{nguyen2022templates} but observed worse performance (Sec.~\ref{sec:abalysis}). We attribute this to the perfect translation assumption made in \cite{nguyen2022templates} which is not the case in our setup.

\section{Experiments}\label{sec:exp}
\subsection{Dataset and Setup}
We consider 5 datasets: LINEMOD~\cite{hinterstoisser2011multimodal}, Occluded-LINEMOD~\cite{brachmann2014learning}, YCB-Video~\cite{xiang2018posecnn}, T-LESS~\cite{hodan2017t}, and YCBInEOAT~\cite{wen2020se}. These involve various challenging scenarios (dense clutter, multi-instance, static or dynamic scenes, table-top or robotic manipulation), and objects with diverse properties (textureless, shiny, symmetric, varying sizes).

As our framework is unified, we consider the combinations among two setups (model-free and model-based) and two pose prediction tasks (6D pose estimation and tracking), resulting in 4 tasks in total. For the model-free setup, a number of reference images capturing the novel object are selected from the training split of the datasets, equipped with the ground-truth annotation of the object pose, following \cite{he2022fs6d}. For the model-based setup, a CAD model is provided for the novel object.  In all evaluation except for ablation, our method always uses the same trained model and configurations for inference \emph{without any fine-tuning}.

\subsection{Metric}\label{sec:metric}
To closely follow the baseline protocols on each setup, we consider the following metrics:
\begin{myitem}
    \item Area under the curve (AUC) of ADD and ADD-S~\cite{xiang2018posecnn}. 
    \item Recall of ADD that is less than 0.1 of the object diameter (ADD-0.1d), as used in \cite{he2022fs6d,he2022oneposepp}.
    \item Average recall (AR) of VSD, MSSD and MSPD metrics introduced in the BOP challenge~\cite{hodan2018bop}.
\end{myitem}

\begin{table}[t]
% \vspace{-0.1in}
\centering
\def\mywidth{0.49\textwidth} 
\definecolor{green}{RGB}{0,160,0}
\resizebox{\mywidth}{!}{

\begin{tabular}{c|cc|cc|cc|cc}
\thickline
                               & \multicolumn{2}{c|}{PREDATOR~\cite{huang2021predator}}          & \multicolumn{2}{c|}{LoFTR~\cite{sun2021loftr}}                  & \multicolumn{2}{c|}{FS6D-DPM~\cite{he2022fs6d}}                 & \multicolumn{2}{c}{\cellcolor[rgb]{ .988,  .894,  .839}Ours} \bigstrut\\
\hline
Ref. images                    & \multicolumn{2}{c|}{16}                                         & \multicolumn{2}{c|}{16}                                         & \multicolumn{2}{c|}{16}                                         & \multicolumn{2}{c}{\cellcolor[rgb]{ .988,  .894,  .839}16} \bigstrut[t]\\
Finetune-free                  & \multicolumn{2}{c|}{\greencheckmark}                            & \multicolumn{2}{c|}{\greencheckmark}                            & \multicolumn{2}{c|}{\redxmark}                                  & \multicolumn{2}{c}{\cellcolor[rgb]{ .988,  .894,  .839}\greencheckmark} \\
Metrics                        & ADD-S                          & ADD                            & ADD-S                          & ADD                            & ADD-S                          & ADD                            & \cellcolor[rgb]{ .988,  .894,  .839}ADD-S & \cellcolor[rgb]{ .988,  .894,  .839}ADD \bigstrut[b]\\
\hline
\multicolumn{1}{l|}{002\_master\_chef\_can} & 73.0                           & 17.4                           & 87.2                           & 50.6                           & 92.6                           & 36.8                           & \cellcolor[rgb]{ .988,  .894,  .839}96.9 & \cellcolor[rgb]{ .988,  .894,  .839}91.3 \bigstrut[t]\\
\multicolumn{1}{l|}{003\_cracker\_box} & 41.7                           & 8.3                            & 71.8                           & 25.5                           & 83.9                           & 24.5                           & \cellcolor[rgb]{ .988,  .894,  .839}97.5 & \cellcolor[rgb]{ .988,  .894,  .839}96.2 \\
\multicolumn{1}{l|}{004\_sugar\_box} & 53.7                           & 15.3                           & 63.9                           & 13.4                           & 95.1                           & 43.9                           & \cellcolor[rgb]{ .988,  .894,  .839}97.5 & \cellcolor[rgb]{ .988,  .894,  .839}87.2 \\
\multicolumn{1}{l|}{005\_tomato\_soup\_can} & 81.2                           & 44.4                           & 77.1                           & 52.9                           & 93.0                           & 54.2                           & \cellcolor[rgb]{ .988,  .894,  .839}97.6 & \cellcolor[rgb]{ .988,  .894,  .839}93.3 \\
\multicolumn{1}{l|}{006\_mustard\_bottle} & 35.5                           & 5.0                            & 84.5                           & 59.0                           & 97.0                           & 71.1                           & \cellcolor[rgb]{ .988,  .894,  .839}98.4 & \cellcolor[rgb]{ .988,  .894,  .839}97.3 \\
\multicolumn{1}{l|}{007\_tuna\_fish\_can} & 78.2                           & 34.2                           & 72.6                           & 55.7                           & 94.5                           & 53.9                           & \cellcolor[rgb]{ .988,  .894,  .839}97.7 & \cellcolor[rgb]{ .988,  .894,  .839}73.7 \\
\multicolumn{1}{l|}{008\_pudding\_box} & 73.5                           & 24.2                           & 86.5                           & 68.1                           & 94.9                           & 79.6                           & \cellcolor[rgb]{ .988,  .894,  .839}98.5 & \cellcolor[rgb]{ .988,  .894,  .839}97.0 \\
\multicolumn{1}{l|}{009\_gelatin\_box} & 81.4                           & 37.5                           & 71.6                           & 45.2                           & 98.3                           & 32.1                           & \cellcolor[rgb]{ .988,  .894,  .839}98.5 & \cellcolor[rgb]{ .988,  .894,  .839}97.3 \\
\multicolumn{1}{l|}{010\_potted\_meat\_can} & 62.0                           & 20.9                           & 67.4                           & 45.1                           & 87.6                           & 54.9                           & \cellcolor[rgb]{ .988,  .894,  .839}96.6 & \cellcolor[rgb]{ .988,  .894,  .839}82.3 \\
\multicolumn{1}{l|}{011\_banana} & 57.7                           & 9.9                            & 24.2                           & 1.6                            & 94.0                           & 69.1                           & \cellcolor[rgb]{ .988,  .894,  .839}98.1 & \cellcolor[rgb]{ .988,  .894,  .839}95.4 \\
\multicolumn{1}{l|}{019\_pitcher\_base} & 83.7                           & 18.1                           & 58.7                           & 22.3                           & 91.1                           & 40.4                           & \cellcolor[rgb]{ .988,  .894,  .839}97.9 & \cellcolor[rgb]{ .988,  .894,  .839}96.6 \\
\multicolumn{1}{l|}{021\_bleach\_cleanser} & 88.3                           & 48.1                           & 36.9                           & 16.7                           & 89.4                           & 44.1                           & \cellcolor[rgb]{ .988,  .894,  .839}97.4 & \cellcolor[rgb]{ .988,  .894,  .839}93.3 \\
\multicolumn{1}{l|}{024\_bowl} & 73.2                           & 17.4                           & 32.7                           & 1.4                            & 74.7                           & 0.9                            & \cellcolor[rgb]{ .988,  .894,  .839}94.9 & \cellcolor[rgb]{ .988,  .894,  .839}89.7 \\
\multicolumn{1}{l|}{025\_mug}  & 84.8                           & 29.5                           & 47.3                           & 23.6                           & 86.5                           & 39.2                           & \cellcolor[rgb]{ .988,  .894,  .839}96.2 & \cellcolor[rgb]{ .988,  .894,  .839}75.8 \\
\multicolumn{1}{l|}{035\_power\_drill} & 60.6                           & 12.3                           & 18.8                           & 1.3                            & 73.0                           & 19.8                           & \cellcolor[rgb]{ .988,  .894,  .839}98.0 & \cellcolor[rgb]{ .988,  .894,  .839}96.3 \\
\multicolumn{1}{l|}{036\_wood\_block} & 70.5                           & 10.0                           & 49.9                           & 1.4                            & 94.7                           & 27.9                           & \cellcolor[rgb]{ .988,  .894,  .839}97.4 & \cellcolor[rgb]{ .988,  .894,  .839}94.7 \\
\multicolumn{1}{l|}{037\_scissors} & 75.5                           & 25.0                           & 32.3                           & 14.6                           & 74.2                           & 27.7                           & \cellcolor[rgb]{ .988,  .894,  .839}97.8 & \cellcolor[rgb]{ .988,  .894,  .839}95.5 \\
\multicolumn{1}{l|}{040\_large\_marker} & 81.8                           & 38.9                           & 20.7                           & 8.4                            & 97.4                           & 74.2                           & \cellcolor[rgb]{ .988,  .894,  .839}98.6 & \cellcolor[rgb]{ .988,  .894,  .839}96.5 \\
\multicolumn{1}{l|}{051\_large\_clamp} & 83.0                           & 34.4                           & 24.1                           & 11.2                           & 82.7                           & 34.7                           & \cellcolor[rgb]{ .988,  .894,  .839}96.9 & \cellcolor[rgb]{ .988,  .894,  .839}92.7 \\
\multicolumn{1}{l|}{052\_extra\_large\_clamp} & 72.9                           & 24.1                           & 15.0                           & 1.8                            & 65.7                           & 10.1                           & \cellcolor[rgb]{ .988,  .894,  .839}97.6 & \cellcolor[rgb]{ .988,  .894,  .839}94.1 \\
\multicolumn{1}{l|}{061\_foam\_brick} & 79.2                           & 35.5                           & 59.4                           & 31.4                           & 95.7                           & 45.8                           & \cellcolor[rgb]{ .988,  .894,  .839}98.1 & \cellcolor[rgb]{ .988,  .894,  .839}93.4 \bigstrut[b]\\
\hline
MEAN                           & 71.0                           & 24.3                           & 52.5                           & 26.2                           & 88.4                           & 42.1                           & \cellcolor[rgb]{ .988,  .894,  .839}\textbf{97.4} & \cellcolor[rgb]{ .988,  .894,  .839}\textbf{91.5} \bigstrut\\

\thickline
\end{tabular}%

}
% \rule{\mywidth}{2pt} 
\vspace{-0.1in}
\caption{Model-free pose estimation results measured by AUC of ADD and ADD-S on YCB-Video dataset. ``Finetuned'' means the method was fine-tuned with group split of object instances on the testing dataset, as introduced by \cite{he2022fs6d}.} \label{tab:ycbv_fewshot}
\vspace{-10pt}
\end{table}

\begin{table*}[h]
\centering
\def\mywidth{0.9\textwidth} 
\definecolor{green}{RGB}{0,160,0}
\resizebox{\mywidth}{!}{

\begin{tabular}{c|ccc|ccccccccccccc|c}
\thickline
\multirow{2}[1]{*}{Method}     & \multirow{2}[1]{*}{Modality}   & Finetune-                      & Ref.                           & \multicolumn{13}{c|}{Objects}                                                                                                                                                                                                                                                                                                                                                                                                              & \multirow{2}[1]{*}{Avg.} \\
                               &                                & free                           & images                         & ape                            & benchwise                      & cam                            & can                            & cat                            & driller                        & duck                           & eggbox                         & glue                           & holepuncher                    & iron                           & lamp                           & phone                          &  \bigstrut[b]\\
\hline
Gen6D~\cite{liu2022gen6d}      & RGB                            & \redxmark                      & 200                            & -                              & 77                             & 66.1                           & -                              & 60.7                           & 67.4                           & 40.5                           & 95.7                           & 87.2                           & -                              & -                              & -                              & -                              & - \bigstrut[t]\\
Gen6D\textsuperscript{*}~\cite{liu2022gen6d} & RGB                            & \greencheckmark                & 200                            & -                              & 62.1                           & 45.6                           & -                              & 40.9                           & 48.8                           & 16.2                           & -                              & -                              & -                              & -                              & -                              & -                              & - \\
OnePose~\cite{sun2022onepose}  & RGB                            & \greencheckmark                & 200                            & 11.8                           & 92.6                           & 88.1                           & 77.2                           & 47.9                           & 74.5                           & 34.2                           & 71.3                           & 37.5                           & 54.9                           & 89.2                           & 87.6                           & 60.6                           & 63.6 \\
OnePose++~\cite{he2022oneposepp} & RGB                            & \greencheckmark                & 200                            & 31.2                           & 97.3                           & 88.0                           & 89.8                           & 70.4                           & 92.5                           & 42.3                           & 99.7                           & 48.0                           & 69.7                           & 97.4                           & 97.8                           & 76.0                           & 76.9 \\
LatentFusion~\cite{park2020latentfusion} & RGBD                           & \greencheckmark                & 16                             & 88.0                           & 92.4                           & 74.4                           & 88.8                           & 94.5                           & 91.7                           & 68.1                           & 96.3                           & 94.9                           & 82.1                           & 74.6                           & 94.7                           & 91.5                           & 87.1 \\
FS6D~\cite{he2022fs6d}         & RGBD                           & \redxmark                      & 16                             & 74.0                           & 86.0                           & 88.5                           & 86.0                           & 98.5                           & 81.0                           & 68.5                           & \textbf{100.0}                 & 99.5                           & 97.0                           & 92.5                           & 85.0                           & 99.0                           & 88.9 \\
FS6D~\cite{he2022fs6d} + ICP   & RGBD                           & \redxmark                      & 16                             & 78.0                           & 88.5                           & 91.0                           & 89.5                           & 97.5                           & 92.0                           & 75.5                           & 99.5                           & 99.5                           & 96.0                           & 87.5                           & 97.0                           & 97.5                           & 91.5 \\
\rowcolor[rgb]{ .988,  .894,  .839} Ours                           & RGBD                           & \greencheckmark                & 16                             & \textbf{99.0}                  & \textbf{100.0}                 & \textbf{100.0}                 & \textbf{100.0}                 & \textbf{100.0}                 & \textbf{100.0}                 & \textbf{99.4}                  & \textbf{100.0}                 & \textbf{100.0}                 & \textbf{99.9}                  & \textbf{100.0}                 & \textbf{100.0}                 & \textbf{100.0}                 & \textbf{99.9} \\

\thickline
\end{tabular}%
}
% \rule{\mywidth}{2pt} 
\vspace{-0.1in}
\caption{Model-free pose estimation results measured by ADD-0.1d on LINEMOD dataset. Gen6D*~\cite{liu2022gen6d} represents the variation without fine-tuning. }
\label{tab:linemod_fewshot}
\vspace{-18pt}
\end{table*}

\subsection{Pose Estimation Comparison}
\boldparagraphstart{Model-free.}  Table~\ref{tab:ycbv_fewshot} presents the comparison results against the state-of-art RGBD methods~\cite{huang2021predator,sun2021loftr,he2022fs6d} on YCB-Video dataset. The baselines results are adopted from \cite{he2022fs6d}. Following \cite{he2022fs6d}, all methods are given the perturbed ground-truth bounding box as 2D detection for fair comparison. Table~\ref{tab:linemod_fewshot} presents the comparison results on LINEMOD dataset. The baseline results are adopted from \cite{he2022fs6d,he2022oneposepp}.  RGB-based methods~\cite{liu2022gen6d,sun2022onepose,he2022oneposepp} are given the privilege of much larger number of reference images to compensate for the lack of depth. Among RGBD methods, FS6D~\cite{he2022fs6d} requires fine-tuning on the target dataset.
Our method significantly outperforms the existing methods on both datasets without fine-tuning on the target dataset or ICP refinement.

\begin{figure}[t]
    \centering
    % \vspace{-0.1in}
    {\includegraphics[width=0.49\textwidth]{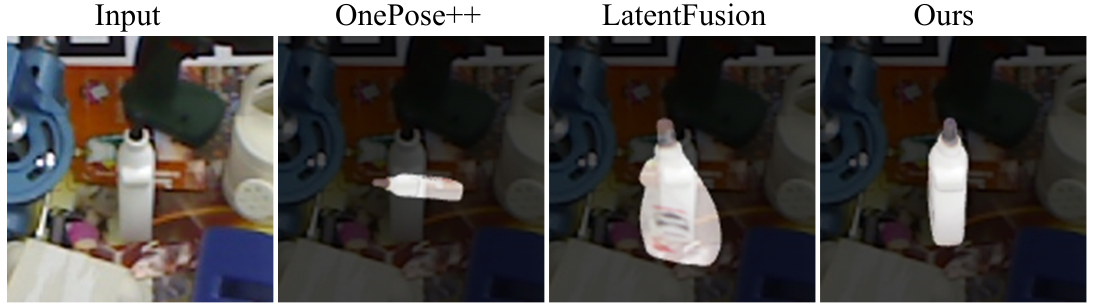}} 
    \vspace{-0.3in}
    \caption{Qualitative comparison of pose estimation on LINEMOD dataset under the model-free setup. Images are cropped and zoomed-in for better visualization.}    \label{fig:linemod_fewshot}
    \vspace{-0.2in}
\end{figure}

Fig.~\ref{fig:linemod_fewshot} visualizes the qualitative comparison. We do not have access to the pose predictions of FS6D~\cite{he2022fs6d} for qualitative results, since its code is not publicly released. The severe self-occlusion and lack of texture on the glue largely  challenge OnePose++~\cite{he2022oneposepp} and LatentFusion~\cite{park2020latentfusion}, while  our method successfully estimates the pose.

\begin{table}[h]
\centering
\def\mywidth{0.45\textwidth} 
\definecolor{green}{RGB}{0,200,0}
\resizebox{\mywidth}{!}{

\begin{tabular}{c|c|ccc|c}
\thickline
\multirow{2}[1]{*}{Method}     & \multicolumn{1}{p{3.555em}|}{Unseen} & \multicolumn{3}{c|}{Dataset}                                                                     & \multirow{2}[1]{*}{ Mean} \\
                               & objects                        & LM-O                           &  T-LESS                        & YCB-V                          &  \bigstrut[b]\\
\hline
SurfEmb~\cite{haugaard2022surfemb} + ICP & \redxmark                      & 75.8                           & 82.8                           & 80.6                           & 79.7 \bigstrut\\
\hline
OSOP~\cite{shugurov2022osop} + ICP & \greencheckmark                & 48.2                           & -                              & 57.2                           & - \bigstrut[t]\\
(PPF, Sift) + Zephyr~\cite{okorn2021zephyr} & \greencheckmark                & 59.8                           & -                              & 51.6                           & - \\
MegaPose-RGBD~\cite{labbemegapose} & \greencheckmark                & 58.3                           & 54.3                           & 63.3                           & 58.6 \\
OVE6D~\cite{cai2022ove6d}      & \greencheckmark                & 49.6                           & 52.3                           & -                              & - \\
GCPose~\cite{zhao2023learning} & \greencheckmark                & 65.2                           & 67.9                           & -                              & - \\
\rowcolor[rgb]{ .988,  .894,  .839} Ours                           & \greencheckmark                & \textbf{78.8}                  & \textbf{83.0}                  & \textbf{88.0}                  & \textbf{83.3} \\

\thickline
\end{tabular}%

}
% \rule{\mywidth}{2pt} 
\vspace{-10pt}
\caption{Model-based pose estimation results measured by AR score on representative BOP datasets. All methods use the RGBD modality.}
\label{tab:bop}
\end{table}

\boldparagraphstart{Model-based.}
Table~\ref{tab:bop} presents the comparison results among RGBD methods on 3 core datasets from BOP: Occluded-LINEMOD~\cite{brachmann2014learning}, YCB-Video~\cite{xiang2018posecnn} and T-LESS~\cite{hodan2017t}. All methods use  Mask~R-CNN~\cite{he2017mask} for 2D detection. Our method outperforms the existing model-based methods that deal with novel objects, and the instance-level method~\cite{haugaard2022surfemb}, by a large margin.

\subsection{Pose Tracking Comparison}\label{exp:track}

\begin{table}[h]
\centering
\def\mywidth{0.5\textwidth} 
\definecolor{green}{RGB}{0,200,0}
\resizebox{\mywidth}{!}{

\begin{tabular}{ll|c|ccccc|c}
\thickline
                               &                                & se(3)-                         & RGF                            & Bundle-                        & Bundle-                        & Wüthrich                       & \cellcolor[rgb]{ .988,  .894,  .839}Ours & \cellcolor[rgb]{ .988,  .894,  .839}Ours\textsuperscript{\textdagger} \bigstrut[t]\\
                               &                                & TrackNet~\cite{wen2020se}      & \cite{issac2016depth}          & Track~\cite{bundle2021wen}     & SDF~\cite{wen2023bundlesdf}    & ~\cite{wuthrich2013probabilistic} & \cellcolor[rgb]{ .988,  .894,  .839} & \cellcolor[rgb]{ .988,  .894,  .839} \bigstrut[b]\\
\hline
\multirow{2}[2]{*}{Properties} & Novel object                   & \redxmark                      & \greencheckmark                & \greencheckmark                & \greencheckmark                & \greencheckmark                & \cellcolor[rgb]{ .988,  .894,  .839}\greencheckmark & \cellcolor[rgb]{ .988,  .894,  .839}\greencheckmark \bigstrut[t]\\
                               & Initial pose                   & \textcolor[rgb]{ 0,  .439,  .753}{GT} & \textcolor[rgb]{ 0,  .439,  .753}{GT} & \textcolor[rgb]{ 0,  .439,  .753}{GT} & \textcolor[rgb]{ 0,  .439,  .753}{GT} & \textcolor[rgb]{ 0,  .439,  .753}{GT} & \cellcolor[rgb]{ .988,  .894,  .839}\textcolor[rgb]{ 0,  .439,  .753}{GT} & \cellcolor[rgb]{ .988,  .894,  .839}\textcolor[rgb]{ .439,  .678,  .278}{Est.} \bigstrut[b]\\
\hline
\multirow{2}[1]{*}{cracker\_box} & ADD-S                          & 94.06                          & 55.44                          & 89.41                          & 90.63                          & 88.13                          & \cellcolor[rgb]{ .988,  .894,  .839}\textbf{95.10} & \cellcolor[rgb]{ .988,  .894,  .839}\textbf{94.92} \bigstrut[t]\\
                               & ADD                            & 90.76                          & 34.78                          & 85.07                          & 85.37                          & 79.00                          & \cellcolor[rgb]{ .988,  .894,  .839}\textbf{91.32} & \cellcolor[rgb]{ .988,  .894,  .839}\textbf{91.54} \\
\multirow{2}[0]{*}{bleach\_cleanser} & ADD-S                          & 94.44                          & 45.03                          & 94.72                          & 94.28                          & 68.96                          & \cellcolor[rgb]{ .988,  .894,  .839}\textbf{95.96} & \cellcolor[rgb]{ .988,  .894,  .839}\textbf{96.36} \\
                               & ADD                            & 89.58                          & 29.40                          & 89.34                          & 87.46                          & 61.47                          & \cellcolor[rgb]{ .988,  .894,  .839}\textbf{91.45} & \cellcolor[rgb]{ .988,  .894,  .839}\textbf{92.63} \\
\multirow{2}[0]{*}{sugar\_box} & ADD-S                          & 94.80                          & 16.87                          & 90.22                          & 93.81                          & 92.75                          & \cellcolor[rgb]{ .988,  .894,  .839}\textbf{96.67} & \cellcolor[rgb]{ .988,  .894,  .839}\textbf{96.61} \\
                               & ADD                            & 92.43                          & 15.82                          & 85.56                          & 88.62                          & 86.78                          & \cellcolor[rgb]{ .988,  .894,  .839}\textbf{94.14} & \cellcolor[rgb]{ .988,  .894,  .839}\textbf{93.96} \\
\multirow{2}[0]{*}{tomato\_soup\_can} & ADD-S                          & 96.95                          & 26.44                          & 95.13                          & 95.24                          & 93.17                          & \cellcolor[rgb]{ .988,  .894,  .839}\textbf{96.58} & \cellcolor[rgb]{ .988,  .894,  .839}\textbf{96.54} \\
                               & ADD                            & 93.40                          & 15.13                          & 86.00                          & 83.10                          & 63.71                          & \cellcolor[rgb]{ .988,  .894,  .839}\textbf{91.71} & \cellcolor[rgb]{ .988,  .894,  .839}\textbf{91.85} \\
\multirow{2}[1]{*}{mustard\_bottle} & ADD-S                          & 97.92                          & 60.17                          & 95.35                          & 95.75                          & 95.31                          & \cellcolor[rgb]{ .988,  .894,  .839}\textbf{97.89} & \cellcolor[rgb]{ .988,  .894,  .839}\textbf{97.77} \\
                               & ADD                            & 97.00                          & 56.49                          & 92.26                          & 89.87                          & 91.31                          & \cellcolor[rgb]{ .988,  .894,  .839}\textbf{96.34} & \cellcolor[rgb]{ .988,  .894,  .839}\textbf{95.95} \bigstrut[b]\\
\hline
\multirow{2}[1]{*}{All}        & ADD-S                          & 95.53                          & 39.90                          & 92.53                          & 93.77                          & 89.18                          & \cellcolor[rgb]{ .988,  .894,  .839}\textbf{96.42} & \cellcolor[rgb]{ .988,  .894,  .839}\textbf{96.40} \bigstrut[t]\\
                               & ADD                            & 92.66                          & 29.98                          & 87.34                          & 86.95                          & 78.28                          & \cellcolor[rgb]{ .988,  .894,  .839}\textbf{93.09} & \cellcolor[rgb]{ .988,  .894,  .839}\textbf{93.22} \\
\thickline
\end{tabular}%

}
% \rule{\mywidth}{2pt} 
\vspace{-0.1in}
\caption{Pose tracking results of RGBD methods measured by AUC of ADD and ADD-S on YCBInEOAT dataset. Ours\textsuperscript{\textdagger} represents our unified pipeline that uses the pose estimation module for pose initialization.}
\label{tab:ycbineoat}
\end{table}

Unless otherwise specified, no re-initialization is applied to the evaluated methods in the case of tracking lost, in order to evaluate long-term tracking robustness. We defer to our supplemental materials for qualitative results.

For comprehensive comparison on the challenges of abrupt out-of-plane rotations, dynamic external occlusions and disentangled camera motions, we evaluate pose tracking methods on the YCBInEOAT~\cite{wen2020se} dataset which includes videos of dynamic robotic manipulation.  Results under the model-based setup are presented in  Table~\ref{tab:ycbineoat}. Our method achieves the best performance and even outperforms the instance-wise training method~\cite{wen2020se} with ground-truth pose initialization. Moreover, our unified framework also allows for end-to-end pose estimation and tracking without external pose initialization, which is the only method with such capability, noted as \textit{Ours\textsuperscript{\textdagger}} in the table.

Table~\ref{tab:ycbv_track} presents the comparison results of pose tracking on YCB-Video~\cite{xiang2018posecnn} dataset. Among the baselines, DeepIM~\cite{li2018deepim}, se(3)-TrackNet~\cite{wen2020se} and PoseRBPF~\cite{deng2019pose} need training on the same object instances, while Wüthrich \emph{et al.}~\cite{wuthrich2013probabilistic}, RGF~\cite{issac2016depth}, ICG~\cite{stoiber2022iterative} and our method can be instantly applied to novel objects when provided with a CAD model.

\begin{table*}[h]
\centering
\def\mywidth{0.95\textwidth} 
\definecolor{green}{RGB}{0,200,0}
\definecolor{blue}{RGB}{31, 119, 180}
\definecolor{orange}{RGB}{173, 23, 89}
\resizebox{\mywidth}{!}{

\begin{tabular}{l||cc|cc|cc||cc|cc|cc|cc||cc}
\thickline
\multicolumn{1}{c||}{\multirow{2}[2]{*}{Approach}} & \multicolumn{2}{c|}{DeeplM~\cite{li2018deepim}}                 & \multicolumn{2}{c|}{se(3)-TrackNet}                             & \multicolumn{2}{c||}{PoseRBPF~\cite{deng2019pose} }             & \multicolumn{2}{c|}{Wüthrich~\cite{wuthrich2013probabilistic}}  & \multicolumn{2}{c|}{RGF~\cite{issac2016depth}}                  & \multicolumn{2}{c|}{ICG~\cite{stoiber2022iterative}}            & \multicolumn{2}{c||}{\cellcolor[rgb]{ .988,  .894,  .839}Ours}  & \multicolumn{2}{c}{\cellcolor[rgb]{ .988,  .894,  .839}Ours\textsuperscript{\dag}} \bigstrut[t]\\
                               &                                &                                & \multicolumn{2}{c|}{~\cite{wen2020se}}                          & \multicolumn{2}{c||}{+ SDF}                                     &                                &                                &                                &                                &                                &                                & \cellcolor[rgb]{ .988,  .894,  .839} & \cellcolor[rgb]{ .988,  .894,  .839} & \cellcolor[rgb]{ .988,  .894,  .839} & \cellcolor[rgb]{ .988,  .894,  .839} \bigstrut[b]\\
\hline
\multicolumn{1}{c||}{Initial pose } & \multicolumn{2}{c|}{GT}                                         & \multicolumn{2}{c|}{GT}                                         & \multicolumn{2}{c||}{PoseCNN}                                   & \multicolumn{2}{c|}{GT}                                         & \multicolumn{2}{c|}{GT}                                         & \multicolumn{2}{c|}{GT}                                         & \multicolumn{2}{c||}{\cellcolor[rgb]{ .988,  .894,  .839}GT}    & \multicolumn{2}{c}{\cellcolor[rgb]{ .988,  .894,  .839}GT} \bigstrut[t]\\
\multicolumn{1}{c||}{Re-initialization} & \multicolumn{2}{c|}{ Yes (290)}                                 & \multicolumn{2}{c|}{No}                                         & \multicolumn{2}{c||}{Yes (2)}                                   & \multicolumn{2}{c|}{No}                                         & \multicolumn{2}{c|}{No}                                         & \multicolumn{2}{c|}{No}                                         & \multicolumn{2}{c||}{\cellcolor[rgb]{ .988,  .894,  .839}No}    & \multicolumn{2}{c}{\cellcolor[rgb]{ .988,  .894,  .839}No} \\
\multicolumn{1}{c||}{Novel object} & \multicolumn{2}{c|}{\redxmark}                                  & \multicolumn{2}{c|}{\redxmark}                                  & \multicolumn{2}{c||}{\redxmark}                                 & \multicolumn{2}{c|}{\greencheckmark}                            & \multicolumn{2}{c|}{\greencheckmark}                            & \multicolumn{2}{c|}{\greencheckmark}                            & \multicolumn{2}{c||}{\cellcolor[rgb]{ .988,  .894,  .839}\greencheckmark} & \multicolumn{2}{c}{\cellcolor[rgb]{ .988,  .894,  .839}\greencheckmark} \\
\multicolumn{1}{c||}{Object setup} & \multicolumn{2}{c|}{\textcolor[rgb]{ 0,  .439,  .753}{Model-based}} & \multicolumn{2}{c|}{\textcolor[rgb]{ 0,  .439,  .753}{Model-based}} & \multicolumn{2}{c||}{\textcolor[rgb]{ 0,  .439,  .753}{Model-based}} & \multicolumn{2}{c|}{\textcolor[rgb]{ 0,  .439,  .753}{Model-based}} & \multicolumn{2}{c|}{\textcolor[rgb]{ 0,  .439,  .753}{Model-based}} & \multicolumn{2}{c|}{\textcolor[rgb]{ 0,  .439,  .753}{Model-based}} & \multicolumn{2}{c||}{\cellcolor[rgb]{ .988,  .894,  .839}\textcolor[rgb]{ 0,  .439,  .753}{Model-based}} & \multicolumn{2}{c}{\cellcolor[rgb]{ .988,  .894,  .839}\textcolor[rgb]{ .439,  .678,  .278}{Model-free}} \\
\multicolumn{1}{c||}{Metric}   & ADD                            & ADD-S                          & ADD                            & ADD-S                          & ADD                            & ADD-S                          & ADD                            & ADD-S                          & ADD                            & ADD-S                          & ADD                            & ADD-S                          & \cellcolor[rgb]{ .988,  .894,  .839}ADD & \cellcolor[rgb]{ .988,  .894,  .839}ADD-S & \cellcolor[rgb]{ .988,  .894,  .839}ADD & \cellcolor[rgb]{ .988,  .894,  .839}ADD-S \bigstrut[b]\\
\hline
002\_master\_chef\_can         & 89.0                           & 93.8                           & 93.9                           & 96.3                           & 89.3                           & 96.7                           & 55.6                           & 90.7                           & 46.2                           & 90.2                           & 66.4                           & 89.7                           & \cellcolor[rgb]{ .988,  .894,  .839}93.6 & \cellcolor[rgb]{ .988,  .894,  .839}97.0 & \cellcolor[rgb]{ .988,  .894,  .839}91.2 & \cellcolor[rgb]{ .988,  .894,  .839}96.9 \bigstrut[t]\\
003\_cracker\_box              & 88.5                           & 93.0                           & 96.5                           & 97.2                           & 96.0                           & 97.1                           & 96.4                           & 97.2                           & 57.0                           & 72.3                           & 82.4                           & 92.1                           & \cellcolor[rgb]{ .988,  .894,  .839}96.9 & \cellcolor[rgb]{ .988,  .894,  .839}97.8 & \cellcolor[rgb]{ .988,  .894,  .839}96.2 & \cellcolor[rgb]{ .988,  .894,  .839}97.5 \\
004\_sugar\_box                & 94.3                           & 96.3                           & 97.6                           & 98.1                           & 94.0                           & 96.4                           & 97.1                           & 97.9                           & 50.4                           & 72.7                           & 96.1                           & 98.4                           & \cellcolor[rgb]{ .988,  .894,  .839}96.9 & \cellcolor[rgb]{ .988,  .894,  .839}98.2 & \cellcolor[rgb]{ .988,  .894,  .839}94.5 & \cellcolor[rgb]{ .988,  .894,  .839}97.4 \\
005\_tomato\_soup\_can         & 89.1                           & 93.2                           & 95.0                           & 97.2                           & 87.2                           & 95.2                           & 64.7                           & 89.5                           & 72.4                           & 91.6                           & 73.2                           & 97.3                           & \cellcolor[rgb]{ .988,  .894,  .839}96.3 & \cellcolor[rgb]{ .988,  .894,  .839}98.1 & \cellcolor[rgb]{ .988,  .894,  .839}94.3 & \cellcolor[rgb]{ .988,  .894,  .839}97.9 \\
006\_mustard\_bottle           & 92.0                           & 95.1                           & 95.8                           & 97.4                           & 98.3                           & 98.5                           & 97.1                           & 98.0                           & 87.7                           & 98.2                           & 96.2                           & 98.4                           & \cellcolor[rgb]{ .988,  .894,  .839}97.3 & \cellcolor[rgb]{ .988,  .894,  .839}98.4 & \cellcolor[rgb]{ .988,  .894,  .839}97.3 & \cellcolor[rgb]{ .988,  .894,  .839}98.5 \\
007\_tuna\_fish\_can           & 92.0                           & 96.4                           & 86.5                           & 91.1                           & 86.8                           & 93.6                           & 69.1                           & 93.3                           & 28.7                           & 52.9                           & 73.2                           & 95.8                           & \cellcolor[rgb]{ .988,  .894,  .839}96.9 & \cellcolor[rgb]{ .988,  .894,  .839}98.5 & \cellcolor[rgb]{ .988,  .894,  .839}84.0 & \cellcolor[rgb]{ .988,  .894,  .839}97.8 \\
008\_pudding\_box              & 80.1                           & 88.3                           & 97.9                           & 98.4                           & 60.9                           & 87.1                           & 96.8                           & 97.9                           & 12.7                           & 18.0                           & 73.8                           & 88.9                           & \cellcolor[rgb]{ .988,  .894,  .839}97.8 & \cellcolor[rgb]{ .988,  .894,  .839}98.5 & \cellcolor[rgb]{ .988,  .894,  .839}96.9 & \cellcolor[rgb]{ .988,  .894,  .839}98.5 \\
009\_gelatin\_box              & 92.0                           & 94.4                           & 97.8                           & 98.4                           & 98.2                           & 98.6                           & 97.5                           & 98.4                           & 49.1                           & 70.7                           & 97.2                           & 98.8                           & \cellcolor[rgb]{ .988,  .894,  .839}97.7 & \cellcolor[rgb]{ .988,  .894,  .839}98.5 & \cellcolor[rgb]{ .988,  .894,  .839}97.6 & \cellcolor[rgb]{ .988,  .894,  .839}98.5 \\
010\_potted\_meat\_can         & 78.0                           & 88.9                           & 77.8                           & 84.2                           & 76.4                           & 83.5                           & 83.7                           & 86.7                           & 44.1                           & 45.6                           & 93.3                           & 97.3                           & \cellcolor[rgb]{ .988,  .894,  .839}95.1 & \cellcolor[rgb]{ .988,  .894,  .839}97.7 & \cellcolor[rgb]{ .988,  .894,  .839}94.8 & \cellcolor[rgb]{ .988,  .894,  .839}97.5 \\
011\_banana                    & 81.0                           & 90.5                           & 94.9                           & 97.2                           & 92.8                           & 97.7                           & 86.3                           & 96.1                           & 93.3                           & 97.7                           & 95.6                           & 98.4                           & \cellcolor[rgb]{ .988,  .894,  .839}96.4 & \cellcolor[rgb]{ .988,  .894,  .839}98.4 & \cellcolor[rgb]{ .988,  .894,  .839}95.6 & \cellcolor[rgb]{ .988,  .894,  .839}98.1 \\
019\_pitcher\_base             & 90.4                           & 94.7                           & 96.8                           & 97.5                           & 97.7                           & 98.1                           & 97.3                           & 97.7                           & 97.9                           & 98.2                           & 97.0                           & 98.8                           & \cellcolor[rgb]{ .988,  .894,  .839}96.7 & \cellcolor[rgb]{ .988,  .894,  .839}98.0 & \cellcolor[rgb]{ .988,  .894,  .839}96.8 & \cellcolor[rgb]{ .988,  .894,  .839}98.0 \\
021\_bleach\_cleanser          & 81.7                           & 90.5                           & 95.9                           & 97.2                           & 95.9                           & 97.0                           & 95.2                           & 97.2                           & 95.9                           & 97.3                           & 92.6                           & 97.5                           & \cellcolor[rgb]{ .988,  .894,  .839}95.5 & \cellcolor[rgb]{ .988,  .894,  .839}97.8 & \cellcolor[rgb]{ .988,  .894,  .839}94.7 & \cellcolor[rgb]{ .988,  .894,  .839}97.5 \\
024\_bowl                      & 38.8                           & 90.6                           & 80.9                           & 94.5                           & 34.0                           & 93.0                           & 30.4                           & 97.2                           & 24.2                           & 82.4                           & 74.4                           & 98.4                           & \cellcolor[rgb]{ .988,  .894,  .839}95.2 & \cellcolor[rgb]{ .988,  .894,  .839}97.6 & \cellcolor[rgb]{ .988,  .894,  .839}90.5 & \cellcolor[rgb]{ .988,  .894,  .839}95.3 \\
025\_mug                       & 83.2                           & 92.0                           & 91.5                           & 96.9                           & 86.9                           & 96.7                           & 83.2                           & 93.3                           & 60.0                           & 71.2                           & 95.6                           & 98.5                           & \cellcolor[rgb]{ .988,  .894,  .839}95.6 & \cellcolor[rgb]{ .988,  .894,  .839}97.9 & \cellcolor[rgb]{ .988,  .894,  .839}91.5 & \cellcolor[rgb]{ .988,  .894,  .839}96.1 \\
035\_power\_drill              & 85.4                           & 92.3                           & 96.4                           & 97.4                           & 97.8                           & 98.2                           & 97.1                           & 97.8                           & 97.9                           & 98.3                           & 96.7                           & 98.5                           & \cellcolor[rgb]{ .988,  .894,  .839}96.9 & \cellcolor[rgb]{ .988,  .894,  .839}98.2 & \cellcolor[rgb]{ .988,  .894,  .839}96.3 & \cellcolor[rgb]{ .988,  .894,  .839}97.9 \\
036\_wood\_block               & 44.3                           & 75.4                           & 95.2                           & 96.7                           & 37.8                           & 93.6                           & 95.5                           & 96.9                           & 45.7                           & 62.5                           & 93.5                           & 97.2                           & \cellcolor[rgb]{ .988,  .894,  .839}93.2 & \cellcolor[rgb]{ .988,  .894,  .839}97.0 & \cellcolor[rgb]{ .988,  .894,  .839}92.9 & \cellcolor[rgb]{ .988,  .894,  .839}97.0 \\
037\_scissors                  & 70.3                           & 84.5                           & 95.7                           & 97s                            & 72.7                           & 85.5                           & 4.2                            & 16.2                           & 20.9                           & 38.6                           & 93.5                           & 97.3                           & \cellcolor[rgb]{ .988,  .894,  .839}94.8 & \cellcolor[rgb]{ .988,  .894,  .839}97.5 & \cellcolor[rgb]{ .988,  .894,  .839}95.5 & \cellcolor[rgb]{ .988,  .894,  .839}97.8 \\
040\_large\_marker             & 80.4                           & 91.2                           & 92.2                           & 96.0                           & 89.2                           & 97.3                           & 35.6                           & 53.0                           & 12.2                           & 18.9                           & 88.5                           & 97.8                           & \cellcolor[rgb]{ .988,  .894,  .839}96.9 & \cellcolor[rgb]{ .988,  .894,  .839}98.6 & \cellcolor[rgb]{ .988,  .894,  .839}96.6 & \cellcolor[rgb]{ .988,  .894,  .839}98.6 \\
051\_large\_clamp              & 73.9                           & 84.1                           & 94.7                           & 96.9                           & 90.1                           & 95.5                           & 61.2                           & 72.3                           & 62.8                           & 80.1                           & 91.8                           & 96.9                           & \cellcolor[rgb]{ .988,  .894,  .839}93.6 & \cellcolor[rgb]{ .988,  .894,  .839}97.3 & \cellcolor[rgb]{ .988,  .894,  .839}92.5 & \cellcolor[rgb]{ .988,  .894,  .839}96.7 \\
052\_extra\_large\_clamp       & 49.3                           & 90.3                           & 91.7                           & 95.8                           & 84.4                           & 94.1                           & 93.7                           & 96.6                           & 67.5                           & 69.7                           & 85.9                           & 94.3                           & \cellcolor[rgb]{ .988,  .894,  .839}94.4 & \cellcolor[rgb]{ .988,  .894,  .839}97.5 & \cellcolor[rgb]{ .988,  .894,  .839}93.4 & \cellcolor[rgb]{ .988,  .894,  .839}97.3 \\
061\_foam\_brick               & 91.6                           & 95.5                           & 93.7                           & 96.7                           & 96.1                           & 98.3                           & 96.8                           & 98.1                           & 70.0                           & 86.5                           & 96.2                           & 98.5                           & \cellcolor[rgb]{ .988,  .894,  .839}97.9 & \cellcolor[rgb]{ .988,  .894,  .839}98.6 & \cellcolor[rgb]{ .988,  .894,  .839}96.8 & \cellcolor[rgb]{ .988,  .894,  .839}98.3 \bigstrut[b]\\
\hline
All Frames                     & 82.3                           & 91.9                           & 93.0                           & 95.7                           & 87.5                           & 95.2                           & 78.0                           & 90.2                           & 59.2                           & 74.3                           & 86.4                           & 96.5                           & \cellcolor[rgb]{ .988,  .894,  .839}\textbf{96.0} & \cellcolor[rgb]{ .988,  .894,  .839}\textbf{97.9} & \cellcolor[rgb]{ .988,  .894,  .839}\textbf{93.7} & \cellcolor[rgb]{ .988,  .894,  .839}\textbf{97.5} \bigstrut\\
\thickline
\end{tabular}%

}
% \rule{\mywidth}{2pt} 
\vspace{-0.1in}
\caption{Pose tracking results of RGBD methods measured by AUC of ADD and ADD-S on YCB-Video dataset. Ours\textsuperscript{\dag} represents our method under the model-free setup with reference images.}\label{tab:ycbv_track}
\vspace{-15pt}
\end{table*}

\subsection{Analysis}\label{sec:abalysis}

\begin{table}[t]
% \vspace{-10pt}
\centering
\def\mywidth{0.35\textwidth} 
\definecolor{green}{RGB}{0,200,0}
\resizebox{\mywidth}{!}{
\begin{tabular}{l|cc}
\thickline
                               & ADD                        & ADD-S \bigstrut\\
\hline
Ours (proposed)                & 91.52                          & 97.40 \bigstrut[t]\\
W/o LLM texture augmentation   & 90.83                          & 97.38 \\
W/o transformer                & 90.77                          & 97.33 \\
W/o hierarchical comparison    & 89.05                          & 96.67 \\
Ours-InfoNCE                   & 89.39                          & 97.29 \bigstrut[b]\\
\thickline
\end{tabular}%

}
% \rule{\mywidth}{2pt} 
\vspace{-0.1in}
\caption{Ablation study of critical design choices.}
\label{tab:ablation}
\vspace{-15pt}
\end{table}

\boldparagraphstart{Ablation Study.} Table~\ref{tab:ablation} presents the ablation study of critical design choices. The results are evaluated by AUC of ADD and ADD-S metrics on the YCB-Video dataset. \textit{Ours (proposed)} is the default version under the model-free (16 reference images) setup. \textit{W/o LLM texture augmentation} removes the LLM-aided texture augmentation for synthetic training. In \textit{W/o transformer}, we replace the transformer-based architecture by convolutional and linear layers while keeping the similar number of parameters. \textit{W/o hierarchical comparison} only compares the rendering and the cropped input trained by pose-conditioned triplet loss (Eq.~\ref{eq:triplet}) without two-level   hierarchical comparison. At test time, it compares each pose hypothesis with the input observation independently and outputs the pose with the highest score. Example qualitative result is shown in Fig.~\ref{fig:pose_rank}. \textit{Ours-InfoNCE} replaces contrast validated pair-wise loss (Eq.~\ref{eq:list_loss}) by the InfoNCE loss as used in \cite{nguyen2022templates}.

\begin{figure}[b]
    \centering
    \vspace{-10pt}
    {\includegraphics[width=0.35\textwidth]{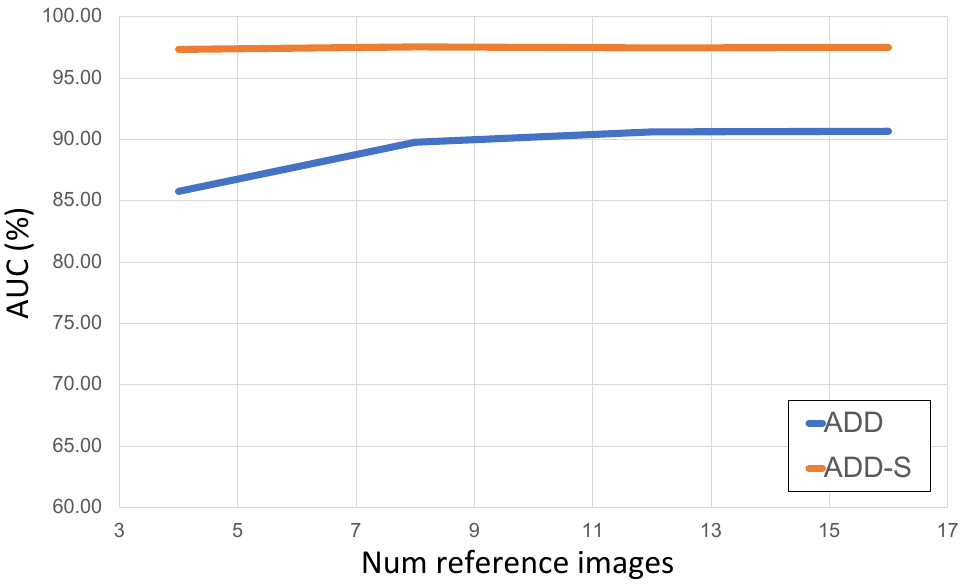}} 
    \vspace{-0.15in}
    \caption{Effects of number of reference images.} \label{fig:num_ref}
    % \vspace{-0.2in}
\end{figure}

\boldparagraphstart{Effects of number of reference images.} We study  how the number of reference images affects the results measured by AUC of ADD and ADD-S on YCB-Video dataset, as shown in Fig.~\ref{fig:num_ref}. Overall, our method is robust to the number of reference images especially on the ADD-S metric, and saturates at 12 images for both metrics. Notably, even when only 4 reference images are provided, our method still yields stronger performance than FS6D~\cite{he2022fs6d} equipped with 16 reference images (Table~\ref{tab:ycbv_fewshot}).

\begin{figure}[h]
    \centering
    {\includegraphics[width=0.35\textwidth]{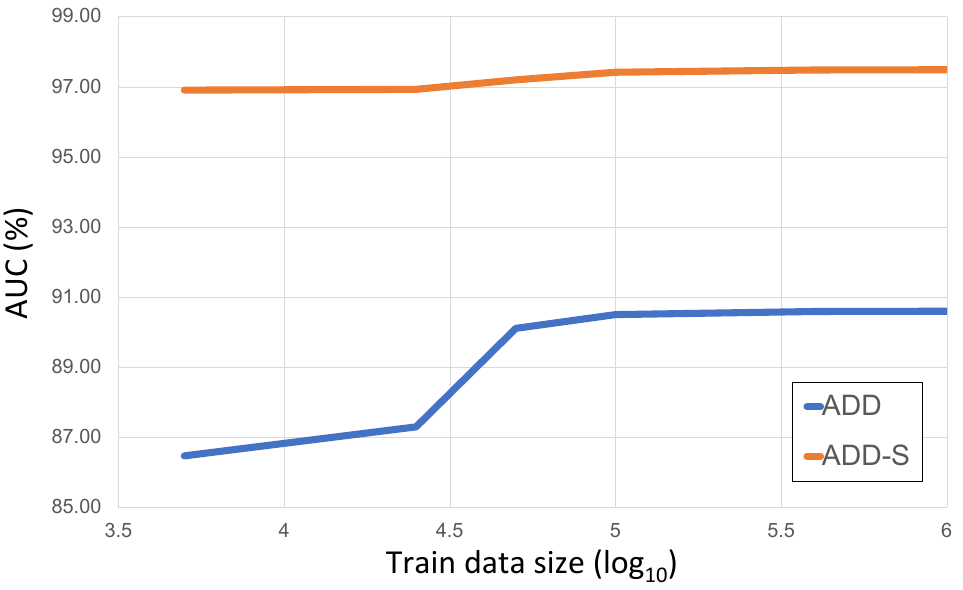}} 
    \vspace{-0.15in}
    \caption{Effects of training data size.} \label{fig:train_size}
    \vspace{-0.2in}
\end{figure}

\boldparagraphstart{Training data scaling law.} Theoretically, an unbounded amount of synthetic data can be produced for training.   Fig.~\ref{fig:train_size} presents how the amount of training data affects the results measured by AUC of ADD and ADD-S metrics on YCB-Video dataset. The gain saturates around 1M.

\boldparagraphstart{Running time.}
We measure the running time on the hardware of Intel i9-10980XE CPU and NVIDIA RTX 3090 GPU. The pose estimation takes about 1.3~s for one object, where pose initialization takes 4~ms, refinement takes 0.88~s, pose selection takes 0.42~s. Tracking runs much faster at $\sim$32~Hz, since only pose refinement is needed and there are not multiple pose hypotheses. In practice, we can run pose estimation once for initialization and switch to tracking mode for real-time performance.

\section{Conclusion}
We present a unified foundation model for 6D pose estimation and tracking of novel objects, supporting both model-based and model-free setups.  Extensive experiments on the combinations of 4 different tasks indicate it is not only versatile but also outperforms existing state-of-art methods specially designed for each task by a considerable margin. It even achieves comparable results to those methods requiring instance-level training. In future work, exploring state estimation beyond single rigid object will be of interest.

% Acknowledgement thank Tianshi Cao.

{
    \small
    \bibliographystyle{ieeenat_fullname}
    \bibliography{ref}
}

\clearpage
\setcounter{page}{1}
\maketitlesupplementary

\begin{figure*}[b]
    \centering
    {\includegraphics[width=0.99\textwidth]{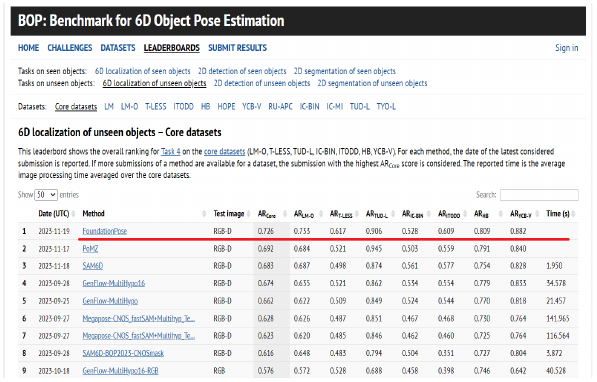}} 
    \vspace{-10pt}
    \caption{Screenshot on BOP leaderboard. At the time of submission, our approach outperforms the previous best method ``PoMZ'' (not yet published) by a considerable margin of 0.03 on AR\textsubscript{Core}, setting a new benchmark record on the leaderboard.}  
    \label{fig:bop}
    \vspace{-10pt}
\end{figure*}

\subsection{Performance on BOP Leaderboard}
Fig.~\ref{fig:bop} presents our results on the BOP challenge of ``6D localization of unseen objects''.\footnote{\url{https://bop.felk.cvut.cz/leaderboards/pose-estimation-unseen-bop23/core-datasets/}} At the time of submission, our FoundationPose is \#1 on the leaderboard.  This corresponds to one of the four tasks considered in this work: model-based pose estimation for novel objects. We use the 2D detection from CNOS~\cite{nguyen2023cnos}, which is the default provided by the BOP challenge.

\subsection{Implementation Details}

During training, for each 3D asset we first pretrain the neural object field with a random number of synthetic reference images. The trained neural object field is then frozen and provides rendering which will be mixed with the model-based OpenGL rendering as input for the pose refinement and selection networks. Such combination better covers the distribution of both model-based and model-free setups, enabling effective generalization as a unified framework.   In terms of the refinement and selection networks, we first train them separately.  We then perform end-to-end fine-tuning for another 5 epochs. The whole training process is conducted over synthetic data  which takes about a week on 4 NVIDIA V100 GPUs. At test time, the model is directly applied to the real world data and runs on one NVIDIA RTX 3090 GPU.  Under the few-shot setup, rendering is obtained from the neural object field which is optimized per object. Under the model-based setup,  rendering is obtained via conventional graphics pipeline~\cite{Laine2020diffrast}. We perform denoising to the depth images implemented in Warp~\cite{warp2022}, which includes erosion and bilateral filtering. The pose-conditioned cropping is implemented in batch using Kornia~\cite{riba2020kornia}.

\boldparagraphstart{Neural Object Field.} We normalize the object into the neural volume bound of $[-1,1]$. The geometry network $\Omega$ consists of two-layer MLP with hidden dimension 64 and ReLU activation except for the last layer. The intermediate geometric feature $f_{\Omega(\cdot)}$ has dimension 16. The appearance network $\Phi$ consists of three-layer MLP with hidden dimension 64 and ReLU activation except for the last layer, where we apply sigmoid activation to map the color prediction to $[0,1]$. We implement the multi-resolution hash encoding~\cite{mueller2022instant} in CUDA and simplify to 4 levels, with number of feature vectors from 16 to 128. Each level's feature dimension is set to 2. The hash table size is set to $2^{22}$. In each iteration the ray batch size is 2048. The truncation distance $\lambda$ is set to 1 cm. In the training loss, $w_e=1, w_s=1000, w_c=100$.  Training takes about 2k steps which is often within seconds.

\begin{figure}[h]
    \centering
    {\includegraphics[width=0.3\textwidth]{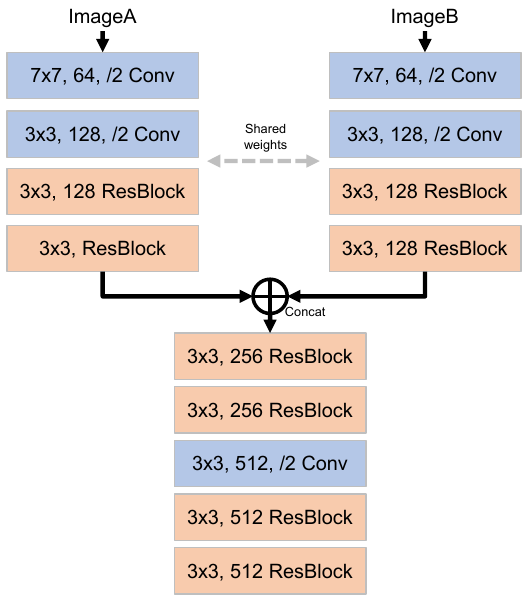}} 
    \vspace{-5pt}
    \caption{Network architecture for image feature embedding used in pose refinement and selection networks. The ResBlock is from ResNet-34~\cite{he2016deep}.}  
    \label{fig:cnn_blcok}
    \vspace{-10pt}
\end{figure}

\boldparagraphstart{Pose Hypothesis Generation.} For global pose initialization, $N_s=42, N_i=12$. To train the refinement network, the pose is randomly perturbed by adding translation noise under the magnitude of $0.02m, 0.02m, 0.05m$ for XYZ axis respectively  and rotation under the magnitude of 20\degree, where the direction is randomized.  Both the rendering and input observation are cropped based on the perturbed pose and resized into 160$\times$160 before sending to the network.  In the training loss (Eq.~\ref{eq:refine_loss}), $w_1$ and $w_2$ are both set to 1. The individual training stage takes 50 epochs. The refinement iteration is set to 1 for training efficiency, At test time, it is set to 5 for pose estimation and 1 for tracking. The complete network architecture of the pose refinement module can be found in the main paper (Fig.~\ref{fig:pipeline}), where the network architecture used for image feature embedding is illustrated in Fig.~\ref{fig:cnn_blcok}. In the transformer encoder, the embedding dimension is 512, number of heads is 4, feed-forward dimension is 512.

\boldparagraphstart{Pose Selection.} The individual training for the selection network  takes 25 epochs, where we perform the similar pose perturbation to refinement network, and the number of pose hypotheses $K=5$. During the end-to-end fine-tuning, the pose hypotheses come from the output of the refinement network. In the training loss (Eq.~\ref{eq:triplet}), $\alpha$ is set to 0.1. The valid positive sample's rotation threshold $d$ is set to 10\degree. The complete network architecture of the pose refinement module can be found in the main paper (Fig.~\ref{fig:pipeline}), where the network architecture used for image feature embedding is illustrated in Fig.~\ref{fig:cnn_blcok}. When performing the two-level hierarchical comparison,  we use the same architecture for both self-attention modules. Concretely, the embedding dimension is 512, number of heads is 4, feed-forward dimension is 512.

\boldparagraphstart{Pose Tracking.} Our framework can be trivially adapted to the pose tracking task while leveraging temporal cues. To do so, at each timestamp, we send the cropped current frame and the rendering using the previous pose to the pose refinement module. The  refined pose becomes the current pose output. This operation repeats along the video sequence. The first frame's pose can be initialized by our pose estimation mode.

\boldparagraphstart{Synthetic Data.} 
 Objaverse assets vary extremely in the object size and mesh complexity. Therefore, we further normalize the objects and remove the disconnected components automatically based on the mesh edge connectivity graph, to make the objects suitable for learning pose estimation.  To create each scene, we randomly sampled $70$ to $90$ objects and dropped them onto a platform with invisible walls until the object velocities were smaller than a threshold. 
We randomly scaled the objects from $5$ to $30$ cm and sampled the size of the platform between $1$ to $1.5$ meter. The LLM-aided texture augmentation is applied to each object from Objaverse~\cite{deitke2023objaverse} with 3 to 5 different seeds for various styles.
To produce diverse and photorealistic images, we randomly created 0 to 5 lights with varied size, color, intensity, temperature and exposure, and $N_c=2$ cameras on a hemisphere with radius ranging from 0.2 to 3.0 meter above the platform. 
We also randomize the material properties, including metallicness and reflection, and textures of the objects and the platform. 
For the environment, we created a dome light with a random orientation and sampled the background from $662$ HDR images obtained from Poly Haven~\cite{polyhaven}. In addition to RGBD rendering, we also store the corresponding object segmentation, camera parameters and the object poses similar to~\cite{labbemegapose, hodan2018bop}. 
In total, our dataset has about 600K scenes and 1.2M images. The dataset will be released on the project page  upon acceptance.

\boldparagraphstart{Creating Reference Images.} In the model-free few-shot setup, similar to \cite{he2022fs6d}, on YCB-Video and LINEMOD datasets, we select a subset of reference images $\mathbb{S}_r$ from the training split $\mathbb{S}_t$. To do so, we first initialize the selection set by choosing the image with the maximum number of pixels according to the mask. Next, for each of the remaining image, we compute its rotational geodesic distance to all the selected reference image, and choose the remaining frame  based on:
\begin{align}
    i^*= \argmaxB_{i \in \mathbb{S}_t, i \notin \mathbb{S}_r} \left( \min_{j \in \mathbb{S}_r} D(\boldsymbol{R}_i, \boldsymbol{R}_j) \right),   
\end{align}
where $D(\cdot, \cdot)$ denotes the geodesic distance on $\mathbb{SO}(3)$. 
We repeat the process until enough number of reference images is obtained, which is typically set to 16 following \cite{he2022fs6d}. 

For applications in the wild when the ground truth object pose is not readily available, we can leverage off-the-shelf SLAM algorithms~\cite{schoenberger2016sfm,teed2021droid,wen2023bundlesdf} to compute the poses from the video. Please refer to our supplemental video for relevant results.

\subsection{Details on Disentangled Representation for Pose Updates.}

\begin{figure}[h]
\vspace{-10pt}
    {\includegraphics[width=0.46\textwidth]{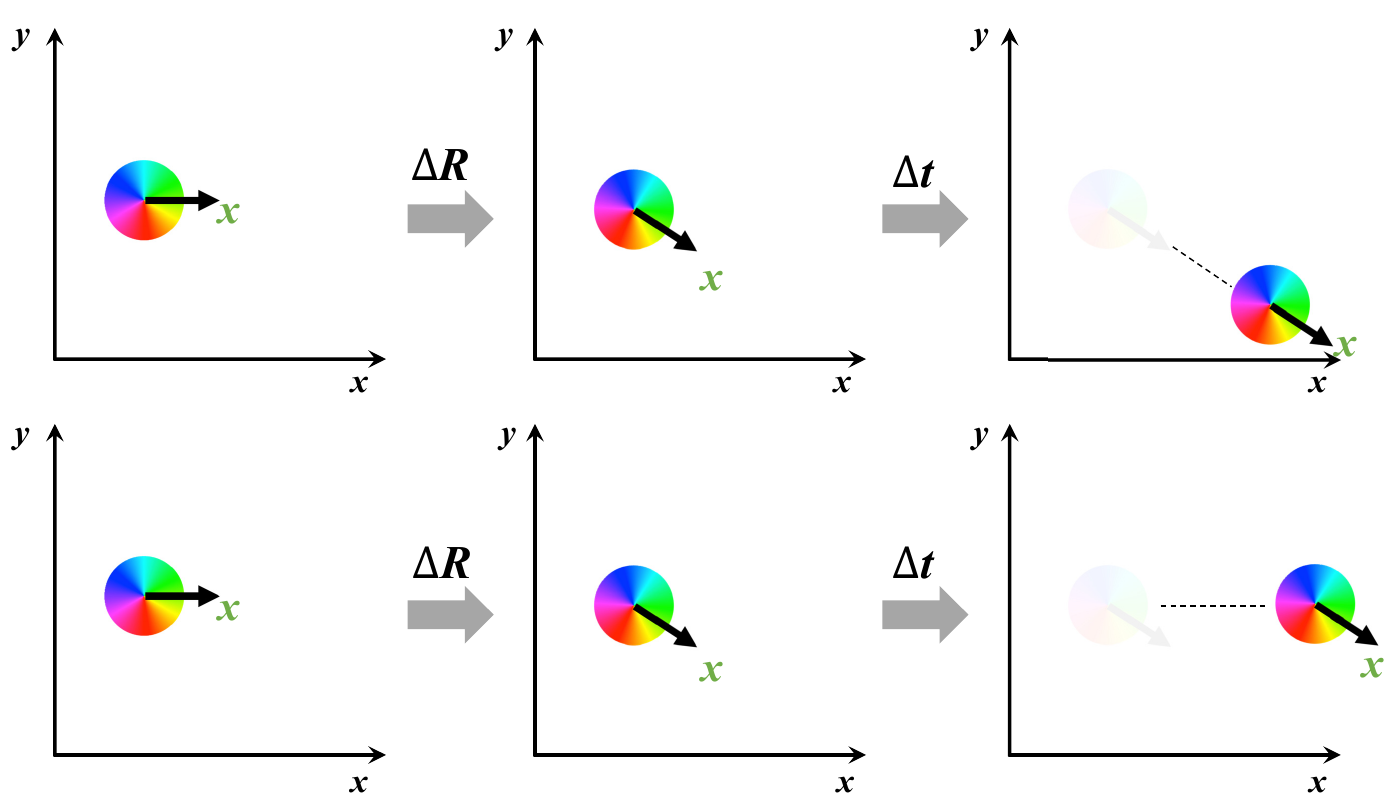}} 
    \vspace{-10pt}
    \caption{Illustration of disentangled representation for pose updates.}\label{fig:disentangle_pose}      
\end{figure}

\noindent As mentioned in the main paper, we disentangle the translation and rotation for two reasons. First,  $\Delta \boldsymbol{t}\in\mathbb{R}^3$ and $\Delta \boldsymbol{R}\in \mathbb{SO}(3)$ are  variables in two different spaces. Therefore, compared to using a single linear projection at the end to predict them jointly, the early disentanglement benefits the learning process. Second, the disentanglement allows us to represent both $\Delta \boldsymbol{t}$ and $\Delta \boldsymbol{R}$ in the camera's coordinate frame, such that $\Delta \boldsymbol{t}$ is independent of $\Delta \boldsymbol{R}$. This is illustrated by a 2D example in Fig.~\ref{fig:disentangle_pose}.
The top row shows the commonly used homogeneous representation, in which the pose update is: $x'=\Delta\boldsymbol{T}x= \Delta\boldsymbol{R}x + \Delta\boldsymbol{t}$.  Thus, $\Delta\boldsymbol{t}$ is applied based on the updated local coordinate system of the disk (object) after applying $\Delta\boldsymbol{R}$, so that the rotation affects the translation. In contrast, the bottom row shows the disentanglement of  $\Delta \boldsymbol{t}$ and  $\Delta \boldsymbol{R}$, which resolves the dependency issue and stabilizes training.

\begin{figure}[h]
    \centering
    {\includegraphics[width=0.45\textwidth]{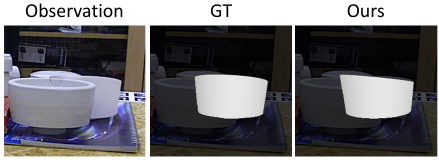}} 
    \vspace{-10pt}
    \caption{Failure mode. Under the combination of multiple challenges including texture-less, severe occlusion, and limited edge cues, our method fails to estimate the correct orientation.}  
    \label{fig:failure}
    % \vspace{-20pt}
\end{figure}

\subsection{Limitations} Similar to related works~\cite{labbemegapose,he2022fs6d,sun2022onepose,he2022oneposepp,cai2022ove6d,zhao2023learning}, our approach focuses on 6D pose estimation and tracking, and relies on external 2D detection, which is obtained from methods such as CNOS~\cite{nguyen2023cnos}, or Mask-RCNN~\cite{he2017mask}. We observe false or missing detection frequently bottlenecks the 6D pose estimation. In future work, an end-to-end framework for novel object detection, 6D pose estimation and tracking would be of interest. Additionally, another typical failure mode due to a combination of multiple challenges is illustrated in Fig.~\ref{fig:failure}.

\subsection{Acknowledgement} 
We would like to thank Tianshi Cao for the valuable discussions; NVIDIA Isaac Sim and Omniverse team for the support on synthetic data generation. 

\end{document}